\documentclass[Afour,sageh,times]{sagej}

\usepackage{moreverb,url}
\usepackage{array}
\usepackage{multirow}
\usepackage{mdwlist}
\usepackage{epsfig} 
\usepackage{epstopdf}
\usepackage{graphics}
\usepackage{graphicx}

\newcommand\BibTeX{{\rmfamily B\kern-.05em \textsc{i\kern-.025em b}\kern-.08em
T\kern-.1667em\lower.7ex\hbox{E}\kern-.125emX}}

\begin{document}

\runninghead{Faraji et al.}

\title{Time-projection control to recover inter-sample disturbances, application to bipedal walking control}

\author{Salman Faraji\affilnum{1}, Philippe M\"{u}llhaupt\affilnum{1} and Auke J. Ijspeert\affilnum{1}}

\affiliation{\affilnum{1}EPFL, Switzerland}

\corrauth{Salman Faraji, EPFL STI IBI-STI BIOROB, Station 9, CH-1015 Lausanne, Switzerland}

\email{salman.faraji@epfl.ch}

\begin{abstract}
We present a new walking controller based on 3LP, a 3D model of bipedal walking that is composed of three pendulums to simulate falling, swing and torso dynamics. Taking advantage of linear equations and closed-form solutions of 3LP, the proposed controller projects intermediate states of the biped back to the beginning of the phase for which a discrete LQR controller is designed. After the projection, a proper control policy is generated by this LQR controller and used at the intermediate time. The projection controller reacts to disturbances immediately and compared to the discrete LQR controller, it provides superior performance in recovering intermittent external pushes. Further analysis of closed-loop eigenvalues and disturbance rejection strength show strong stabilization properties for this architecture. An analysis of viable regions also show that the proposed controller covers most of the maximal viable set of states. It is computationally much faster than Model Predictive Controllers (MPC) and yet optimal over an infinite horizon.
\end{abstract}


\maketitle


 \section{Introduction}
 
 Performing bipedal locomotion for humanoid robots is a challenging task regarding many different aspects. On one hand, the hardware should be powerful enough to handle the weight and fast motions of the swing legs. On the other hand, controllers require accurate perception and actuation capabilities for better stabilization and predictability of the system. Power and precision are two different and sometimes conflicting requirements. From the perspective of geometry also, complex chains in each leg of the robot make the control problem multi-dimensional. Due to an under-actuated nature and geometric complexities, controlling humanoids in dynamic walking is not trivial. The hybrid nature makes such control paradigm even more complicated since the system model changes in each phase of the motion (i.e., left and right support phases). 
 
 In a model-based controller which benefits from mechanical models of the system, it is common to break the complexity into multiple levels. Actuator dynamics are identified and compensated for in a separate block of position or torque controller. The kinematic or rigid body dynamic model is placed in an interface block to translate Cartesian variables to joint variables and vice versa. Then, the whole complex system is approximated with a simpler template model to describe walking dynamics \cite{kajita1991study, faraji20173lp}. The fewer dimensions of such a template model compared to the actual system makes it possible to apply classical control theories in real time, within capabilities of computation units and agility of the motion. The under-actuated hybrid nature of walking is then handled with template models. In this regard, hierarchical model-based control approaches can handle complexities in different levels and use models to capture main dynamics of the actuators, limbs or the full body. 
 
 \subsection{Template models}
 
 Simplified models can speed up the calculation of footstep plans, Center of Mass (CoM) and foot Center of Pressure (CoP) trajectories. A good walking controller should stabilize the under-actuated part of the system (also called falling dynamics) either by modulating the CoP through ankle torques, or by regulating the angular momentum through rotation of the torso, or finally by taking proper footsteps (all discussed in \cite{stephens2007humanoid}). The first two strategies provide continuous control, but with strict bounds on the available ankle torques (determined by the foot size) or body rotations. The third strategy, however, uses the hybrid nature of walking and stabilizes the system in successive phases by adjusting footstep locations. Inverted Pendulum (IP) \cite{kuo2005energetic} and its linear version (LIP) \cite{kajita1991study, capturability} are probably the simplest models used for these three strategies, concentrating the whole mass of the robot in a point and modeling the legs with massless inverted pendulums. In these models, swing dynamics is absent, and therefore, the timing and the final swing-leg attack angle is imposed by the controller. A proper attack angle which directly translates to footstep location can stabilize the system through the third strategy, but it might require a precise full-model inverse dynamics to ensure CoM and swing motion tracking. There are more complex versions of inverted pendulum with masses in the legs \cite{byl2008approximate}, torso \cite{westervelt2007feedback} and knee for the swing leg \cite{asano2004novel}. In all advanced versions of IP, due to non-linearity, optimizations and numerical integrations are needed to obtain periodic walking gaits. These gaits are indeed more natural than simple IP-based gaits, due to inclusion of swing/torso dynamics.
 
 \subsection{Discrete control}
 
\begin{figure}[]
     \centering
     \includegraphics[trim = 1mm 1mm 1mm 1mm, clip, width=0.5\textwidth]{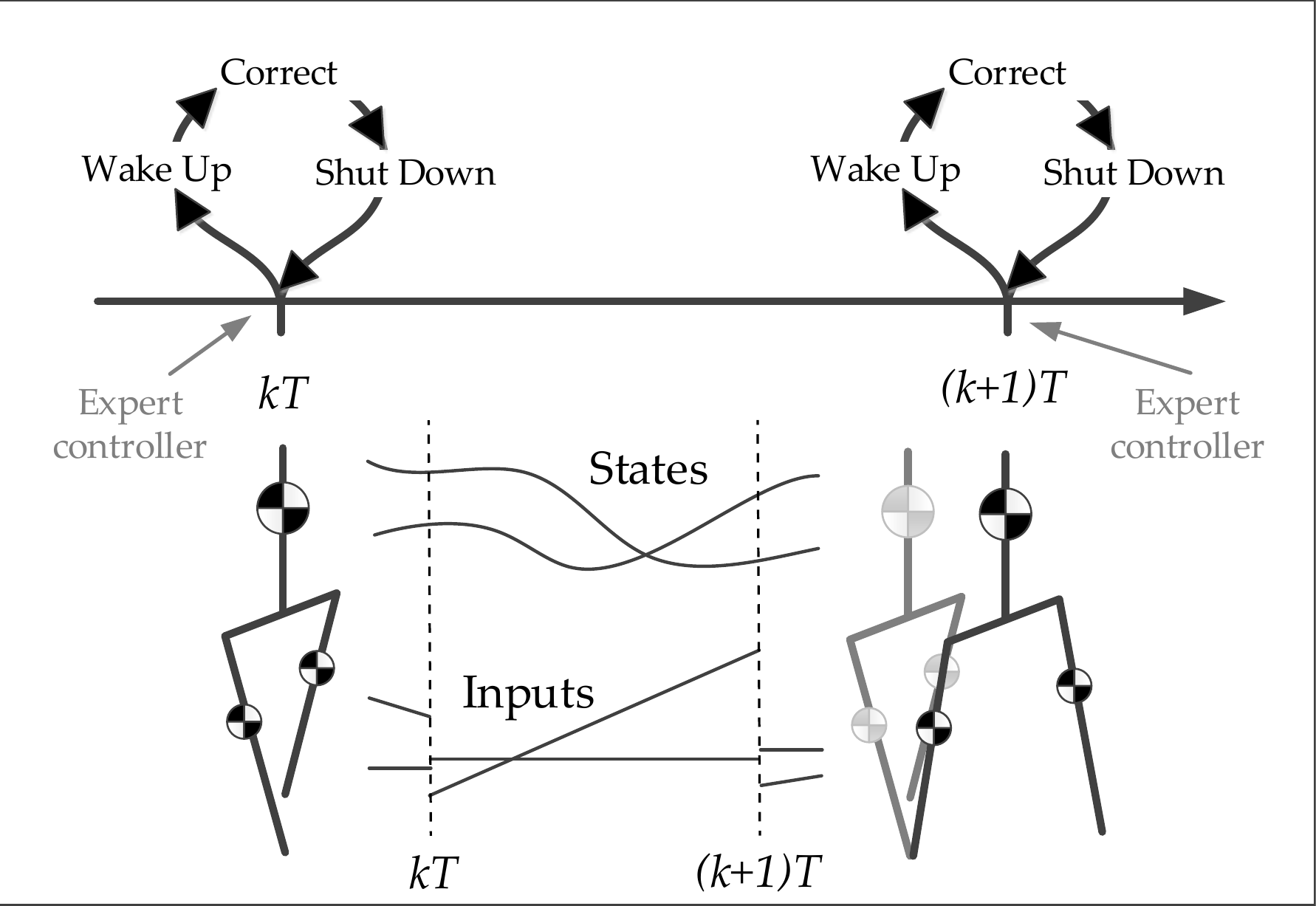}
     \caption{The role of a discrete controller in regulating inputs at the beginning of each phase to stabilize the robot and to track the nominal periodic solution. After one reaction, this controller has to wait until the next event to correct for the accumulated error of all inter-sample disturbances.} 
     \label{fig::discrete_controller}
 \end{figure}

 Linearization of template systems around periodic walking gaits offers a linear model for control and stabilization using the third stabilization strategy \cite{rummel2010stable}. Such a linear model describes the evolution of possible deviations from nominal trajectories and the effect of inputs, particularly footstep locations. Discretization of this linear model between specific gait events, e.g., touch down or maximum apex moments can form the basis for a discrete controller that adjusts inputs when the event triggers. This approach is shown in Figure \ref{fig::discrete_controller} where an expert controller provides corrective inputs based on deviations observed at each event. The linear inverted pendulum (LIP) provides analytical solutions in this regard and is widely used in slow-walking locomotions \cite{feng20133d, faraji2014robust, herdt2010walking, capturability}. In this model, similar to IP, the next footstep location and timing should be imposed as the original model does not include swing dynamics.
 
 \subsection{Limitations of discrete control}
 
 The discrete control paradigm of Figure \ref{fig::discrete_controller} benefits from a very simple discrete map and the control rate is adjusted to the frequency of phase-switches. The synchrony of control with the hybrid phase-changes is of particular interest in walking. In the frame of third stabilization strategy through footstep locations, once a swing phase starts, the falling dynamics keeps growing until the next contact point is established to absorb the extra energy. An effective stabilizing correction is therefore applied only at hybrid phase-change moments \cite{zaytsev2015two, kelly2015non, byl2008approximate}. The first strategy (CoP modulation) offers continuous stabilization authority through modulation of CoP, though the effect is insufficient due to small foot dimensions. This strategy seems more suitable for slow walking speeds and short footsteps, popular for controlling humanoid robots. The second strategy (angular momentum) might lead to unwanted torso oscillations in practice and can not be used alone unless a power-full momentum wheel is mounted on the robot. The third strategy (foot stepping) is more effective in faster speeds and larger footsteps which explains the synchrony and motivates for discretization of the linearized model at hybrid phase-changes, but one should create a library of controllers to handle linearization around different gait conditions \cite{kelly2015non, manchester2014real, gregg2012control}.
 
\begin{figure}[]
     \centering
     \includegraphics[trim = 0mm 7mm 0mm 0mm, clip, width=0.5\textwidth]{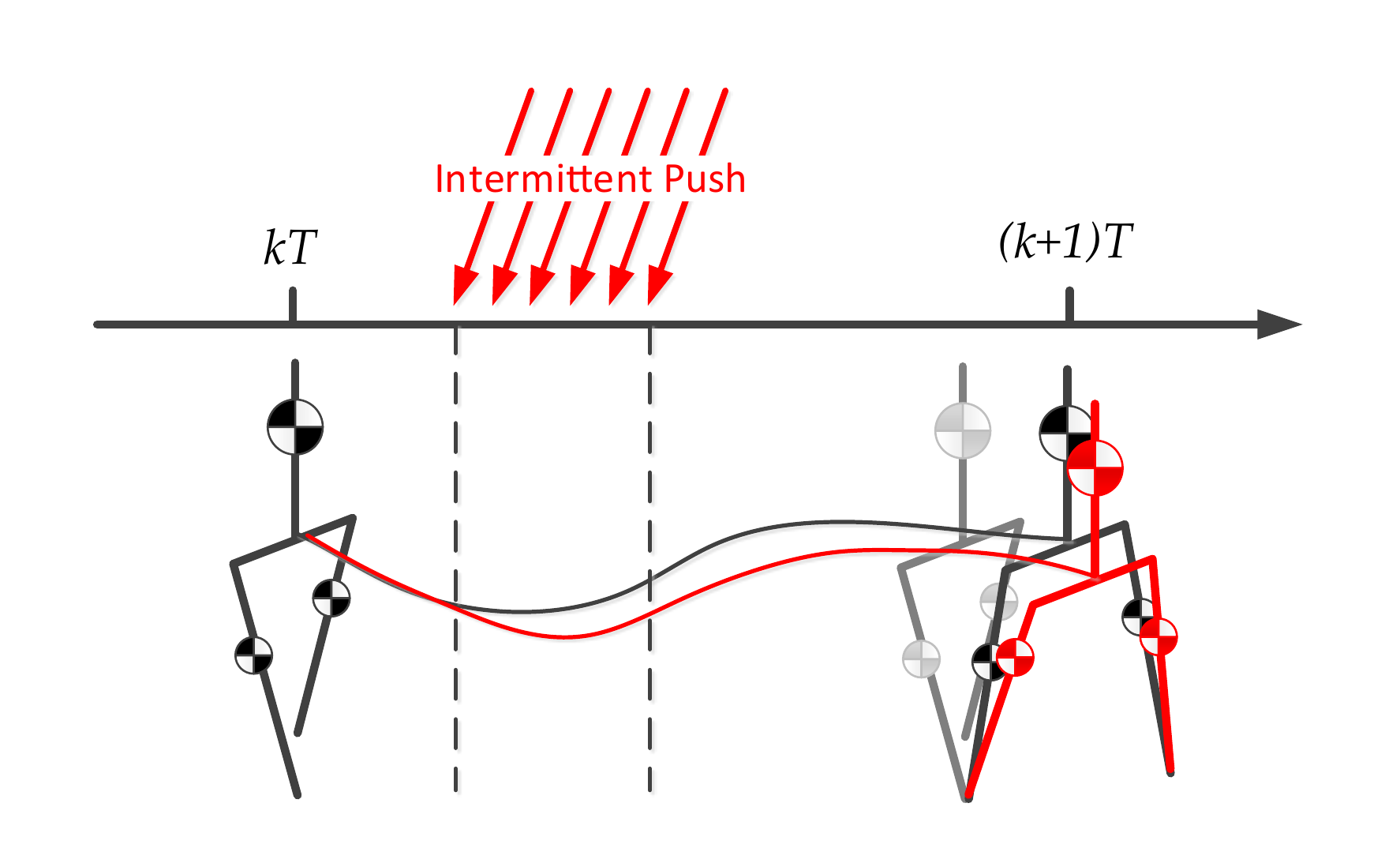}
     \caption{Demonstration of an intermittent push that appears shortly during a continuous phase and influences the system. The normal and disturbed trajectories are shown in black and red respectively. Here, the final footstep location of this phase is not adjusted online. A delayed reaction to disturbance only at the beginning of the next phase might produce a large overshoot in the next footstep locations.} 
     \label{fig::intermittent_push}
 \end{figure}

 Although a discrete model can predict the future very rapidly in terms of  computation, there might be intermittent disturbances that shortly act on the system at any time and disappear. As mentioned earlier, discretization of walking models is usually synchronized with phase-change moments. However, due to unstable falling dynamics, the phase period (i.e., stance period) is long enough for even moderate intermittent disturbances to accumulate and result in a considerable deviation. Such destructive effect (shown in Figure \ref{fig::intermittent_push}) requires taking a large step in the next phase whereas the footstep location of that same phase could be adjusted online to stabilize with less effort. This raises an interesting control problem in which the discretization rate is reasonable but too slow to handle inter-sample disturbances. 
 
 \subsection{Continuous control}
 
 As mentioned earlier, the next footstep has a vital stabilization role, but effective only once the next contact is established. The interesting control problem is in online adjustment of this footstep location, although it does not have a considerable effect instantaneously in the middle of the phase. Without such online adjustment and keeping the swing destination fixed, the second next footstep (which is found in the next event) might become very large to compensate for the accumulated error. This problem might naturally happen in a hybrid system where a deviation of one state (e.g., swing location) might not be important in one phase due to a weak mechanical coupling, but becomes very important in the next phase where it has a strong mechanical coupling (e.g., when it becomes stance foot location). This effect is due to a hybrid change of the system model through motion phases. In normal continuous systems without hybrid phases, however, the coupling between variables remains the same. In other words, since the continuous model remains the same, it is possible to discretize at any rate and apply standard controllers such as Discrete Linear Quadratic Regulators (DLQR) \cite{ogata1995discrete}. If inter-sample disturbances were found to be significant for these normal systems, one could naturally increase the control rate. In our hybrid system, however, this is not straightforward and leads to a complicated model-variant DLQR design. Therefore, DLQR controllers are not used very often for online control of footstep adjustment, although they are extensively used in the first stabilization strategy (CoP modulation) and traditional discrete Poincar\'e based methods \cite{poincare} which linearize the whole motion phase. 
 
 Despite complications of DLQR design in hybrid systems such as walking robots, there are alternative ways to realize online adjustments with very simple laws. Raibert \cite{raibert1984experiments} used a known yet simple approach for hopping where the footstep adjustment was a function of CoM forward speed. This intuitive law with a hand-tuned coefficient moved the footstep location further when a faster forward speed was detected. This idea was later re-formulated for walking with the linear inverted pendulum in \cite{stephens2007humanoid, capturability} where the coefficient had a more physical meaning. The idea was to capture the motion, i.e., to find a footstep location where the CoM ends up on top of this point with zero velocity. In both frameworks, the footstep location can be adjusted online in a fast reaction to intermittent pushes and disturbances. 
 
 In \cite{faraji2014robust}, we used analytical solutions of LIP and a Model Predictive Controller (MPC) to predict CoM trajectories and adjust footstep locations. At each instance of time, we used closed-form solutions of LIP to predict state evolution until the end of the same phase as well as few successive phases. Considering footstep locations as inputs for this discrete system, MPC was able to adjust the next footsteps to stabilize the robot, recover from pushes and converge to a specific walking gait. This numerical optimization could extend the capturability framework \cite{capturability} while providing the same online adjustment capabilities. We were also able to generate lateral bounces and turning gaits naturally inside the optimization while the capturability framework does not have this flexibility. The idea of capture point helps stabilizing the robot when the ankle torques are not enough, i.e. when the instantaneous capture point falls outside the support polygon. The capturability framework, however, requires an intuitive proportional gain to move the desired capture point away from the robot when a walking behavior is desired \cite{capturability2}. Besides, swing trajectories are artificially designed to reach the destination, because the underlying template model (LIP) does not have swing dynamics. In contrary, our proposed controller automatically generates walking gaits and swing motions while the stabilizing controller is designed for both in-place and progressing walking gaits. However, inequality constraints of \cite{capturability} are not yet considered in our method which would be discussed later in the section of viable regions.
 
 \subsection{A better template model?}
 
 In our recent work mentioned earlier \cite{faraji2014robust}, the resulting Cartesian CoM trajectory and an artificial swing trajectory (which terminates in the optimized footstep location) are given to an inverse dynamics interface to find required joint torques for the robot. Although MPC provided a robust performance by finding proper footstep locations \cite{faraji2014robust}, the practical and feasible range of motions strictly depended on the choice of timing, artificial swing trajectories, foot dimensions and the maximum step-length which should not violate the constant CoM height assumption. This motivated us to develop a more complicated yet linear model that captures falling, swing and torso dynamics altogether. This model, called 3LP \cite{faraji20173lp} is composed of three pendulums for the torso and the two legs with masses that all remain in constant-height planes. Closed-form solutions make 3LP suitable for online MPC similar to the framework we introduced in \cite{faraji2014robust}. In this regard, 3LP could relieve the need to produce artificial swing trajectories added to LIP. The inclusion of swing dynamics, however, changed the gait generation paradigm completely. 
 
 Remember that with the LIP model, the controller has to impose the attack angle and phase time. This can create a library of gaits with different speeds and frequencies, but not necessarily feasible to track with inverse dynamics. According to early bio-mechanics studies (e.g., \cite{zarrugh1978predicting}), human's walking frequency increases with speed due to certain mechanical and muscle properties of the body. The particular ascending relation observed in human is the result of optimizing Cost of Transport (CoT) \cite{bertram2005constrained} defined as the energy spent over the unit of traveled distance. This quantity is due to a trade-off between falling, swing and leg lift dynamics. When translated to humanoids, the concept approximately holds in the sense that the ankle, knee and hip joints have to realize push-off, leg lift, and swing works. Thanks to the inclusion of swing dynamics, 3LP uses joint torques to describe the swing motion \cite{faraji20173lp} while LIP is blind to joint torques entirely. Gait symmetry concepts applied to the closed-form equations of 3LP can produce a class of gaits which could be optimized energetically, based on their required hip torques. Such optimal gaits have some dynamic torque and ground reaction profiles similar to the human, despite the constant CoM height assumption.
 
 3LP gaits are obtained by combining eigenvectors of a particular matrix linearly \cite{faraji20173lp} without a numerical integration. Thanks to linear equations, the linearization of system equations around these gaits is very easy to calculate. The tracking problems of inverse dynamics arising with the choice of gait timing and artificial swing trajectories with LIP are now improved with dynamics of 3LP, and finally, we can still find analytic transition matrices between any two instances of time in the phase with 3LP. With these new features, is MPC still needed? As we will see next, the answer is no.
 
 \subsection{Time-projection controller}
 
 \begin{figure*}[]
     \centering
     \includegraphics[trim = 0mm 0mm 0mm 0mm, clip, width=1\textwidth]{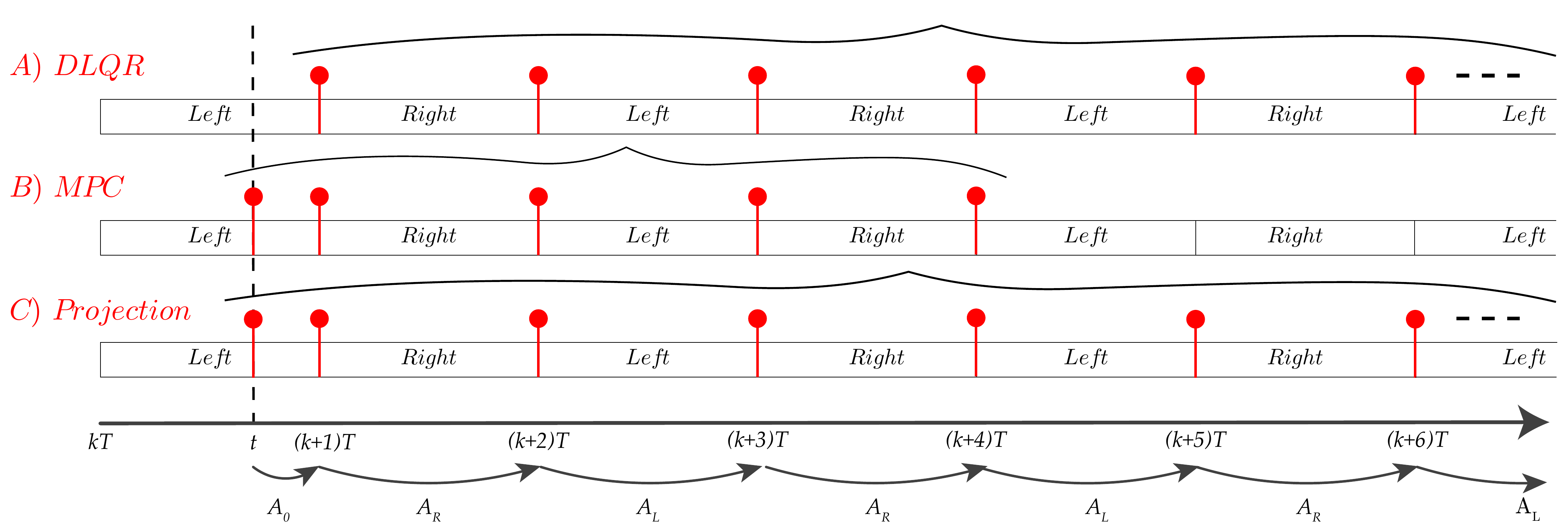}
     \caption{Conceptual comparison of A) DLQR controller, B) receding horizon MPC and C) the proposed projection controller in dealing with the state observed at current inter-sample time $t$ in the left support phase. The DLQR controller only updates inputs at certain discrete events. The receding-horizon MPC controller we used in \cite{faraji2014robust} used the remaining-phase transition matrix $A_0$ and the next transition matrices $A_L, A_R$ to optimize the footsteps at each time-step and provide online reaction. The proposed projection controller provides the same online updating feature, but uses the same DLQR expertise without any numerical optimization of MPC.} 
     \label{fig::projecting-vs-mpc}
 \end{figure*}
 
 Intermittent disturbances can have different magnitude, duration and timing. In walking, early-phase pushes can disturb the system more severely due to the unstable nature of falling dynamics. Traditional Poincar\'e based methods \cite{poincare} which linearize the system around a pre-optimized gait cannot capture such continuous effects. In other words, all disturbances happening between the two discrete events are accumulated and observed only at the next event. To set up a continuous control, however, there should be a measure to evaluate the error in the middle of a continuous phase. Also, one also needs to know the effect of available continuous inputs (like hip torques/footstep locations). Any-time transition matrices of LIP and 3LP make this evolution prediction possible. 
 
 Although LIP can produce gaits used for DLQR control (i.e. event to event control of Figure \ref{fig::discrete_controller}), we used MPC in \cite{faraji2014robust} to handle intermittent pushes. The difference between these two frameworks can be seen in Figure \ref{fig::projecting-vs-mpc}. A DLQR controller could be used on top of LIP gaits to find discrete adjustments (Figure \ref{fig::projecting-vs-mpc}.A). However, when performing continuous online control, we needed MPC to handle the remaining-phase transition matrix $A_0$ (in Figure \ref{fig::projecting-vs-mpc}) and the hybrid nature of phase-switches in a receding horizon fashion (Figure \ref{fig::projecting-vs-mpc}.B). An infinite-horizon MPC without constraints translates to DLQR, but numerical optimizations of MPC with limited horizons offered flexibility to model switches and the inclusion of inequality constraints \cite{faraji2014robust}. 
 
 In this work, we propose a new online control approach based on the DLQR controller. In discrete control architectures \cite{kelly2015non, byl2008approximate, manchester2014real, sharbafi2012controllers}, a certain event is considered to decide the new angle of attack or push-off force. Similarly, we use the DLQR controller in our proposed architecture as a core stabilizing expert. For any time instance $t$ in the middle of a continuous phase, we project or map the currently measured error back in time to the previous touch-down event, where the DLQR controller knows best how to handle it. We take the output of DLQR then and apply it to the system in the middle of the phase at time $t$. Such online policy makes sure that the future evolution of the system given the calculated input will be optimal if seen in a longer time-span over multiple future steps (Figure \ref{fig::projecting-vs-mpc}.C). Because of unstable falling dynamics, a push of certain magnitude and duration might have different effects if applied early or late during single support. Our continuous projection controller is therefore expected to handle such sensitivity to timing, i.e., handling continuous disturbances.
 
\begin{figure}[]
     \centering
     \includegraphics[trim = 3mm 0mm 3mm 0mm, clip, width=0.5\textwidth]{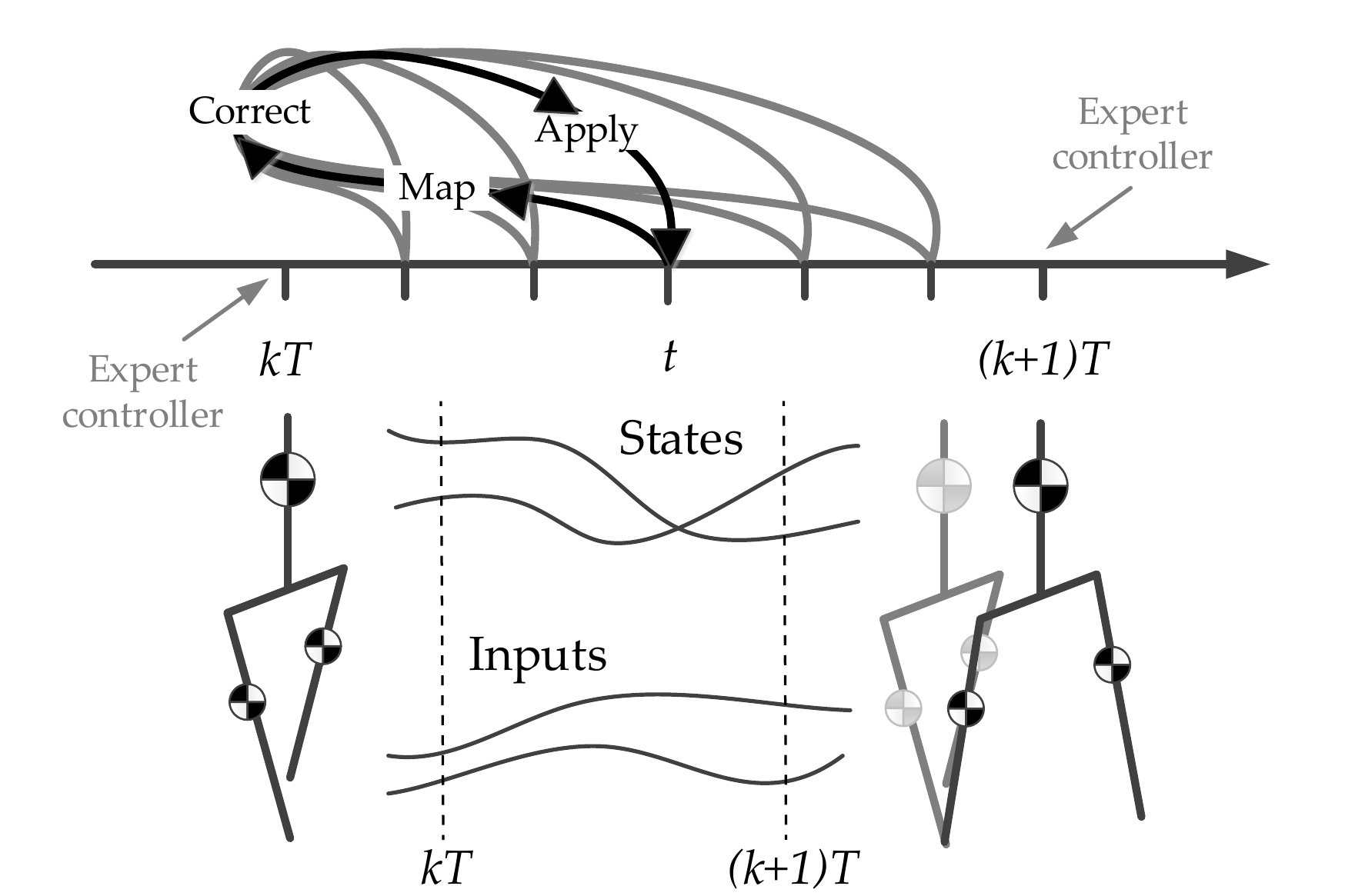}
     \caption{A demonstrative schematic of time-projection idea. This controller in fact relies on the expertise of the DLQR controller. At each time-step $t$, the measured state is mapped to the beginning of the phase, optimal inputs are calculated using the DLQR controller and then, these inputs are used at the current time-step $t$. } 
     \label{fig::cont_controller}
 \end{figure}

 Figure \ref{fig::cont_controller} demonstrates the time-projection idea. This new method is simple to execute, and it does not need a redesign of the controller, off-line optimizations or numerical MPC optimizations. Back projection idea is also used in \cite{byl2008approximate} where a post-collision state should be rewound to a pre-collision state where an additional impulse should be added. Then the system is forward simulated again to find the contribution of the additional impulse. Here we use a similar projection to map the continuous error at any time back to the discrete event, but then use the LQR controller to find control inputs to be used at the current time-step which is conceptually different.
 
 \subsection{Novelty}
 
 Time-projecting control for a walking system is based on three main motivations:
 \begin{enumerate}
     \item DLQR control rate is normally synchronized with the walking frequency. Due to the unstable falling dynamics, this controller is sensitive to the timing of inter-sample disturbances.
     \item The DLQR controller can already stabilize the system through footstep adjustments. There is no need to redesign this part of the controller to capture inter-sample disturbances.
     \item An online controller is needed. We want to benefit from DLQR's stabilization property, but avoid numerical optimizations of a receding horizon MPC.
 \end{enumerate}
 The contribution of this work lies in formulating the time-projection idea and applying it to walking with 3LP in simulations. The aim is to achieve the same performance of online MPC \cite{faraji2014robust,feng20133d, kuindersma2014efficiently} in push recovery scenarios, but without numerical optimizations which might need a considerable computational power. We use 3LP for generation of more feasible gaits compared to LIP, although all control concepts introduced could be applied to LIP as well. An advantage of MPC could be the incorporation of inequality constraints on input torques, the center of pressure, friction cones and footstep length to ensure feasibility of the plan. The proposed method does not consider inequality constraints, but thanks to an analysis of viable regions \cite{zaytsev2015two}, we provide a simple criterion that indicates whether projection controller fails or not, i.e., whether the robot should switch to a more complex controller or even emergency cases. Further analysis of viable regions indicates that extreme conditions rarely happen in normal frequencies and stride lengths. Therefore, our simple projection controller is enough most of the time, and if not, all other controllers including MPC can hardly improve stabilization. We would also like to remark that 3LP already describes the pelvis width and thus avoids internal collisions in natural walking. In extreme cases where a lateral crossover is needed, an MPC controller with non-convex constraints or advanced collision avoidance algorithm might be needed. 
 
 The rest of the article starts with formulation and discussion of time-projection idea for general linear systems. We briefly introduce the 3LP model then, including formulations required for the projection control. Demonstration of intermittent push recovery and speed tracking scenarios follow afterward with an extensive analysis of system eigenvalues, push recovery strength and viable regions. We conclude the paper by comparing the proposed controller with other candidate controllers in the literature and discuss broader control applications which could benefit from the time-projection idea.


\section{Time-Projection}

Continuous-time LQR controllers aim at optimizing system performance over an infinite horizon in future. In discretized and digitized controllers, this translates to a DLQR controller with similar discrete system evolution formulas. We mention that for walking applications, the discretization rate is often set to the step frequency. This is mainly because over the hybrid model switch, the final swing location which weakly couples with other system variables (i.e., CoM state) in the swing phase will have a strong coupling in the stance phase. Since footstep adjustment becomes effective only when the next contact is established, the actual control update events are set to the start of hybrid phases in discrete controllers. We challenged this paradigm by inter-sample pushes and motivated the necessity of updating footstep locations online, in the middle of the swing phase. At the same time, we want to keep the same optimality of DLQR controllers over an infinite horizon.

Time-projection is more suitable for systems that have hybrid phases with a variable model in each phase. In ordinary systems with fixed models and continuous phases, a simple increase in discretization frequency is more straightforward and probably more effective. We explain the time-projection idea with continuous models though, aiming at simpler and more comprehensible formulations. The 3LP model and hybrid phase-changes are discussed later in the 3LP section, and appropriate time-projection formulas are included in the appendix for further information.  

\subsection{Linear System}

\begin{figure*}[]
	\centering
	\includegraphics[trim = 0mm 0mm 0mm 0mm, clip, width=1\textwidth]{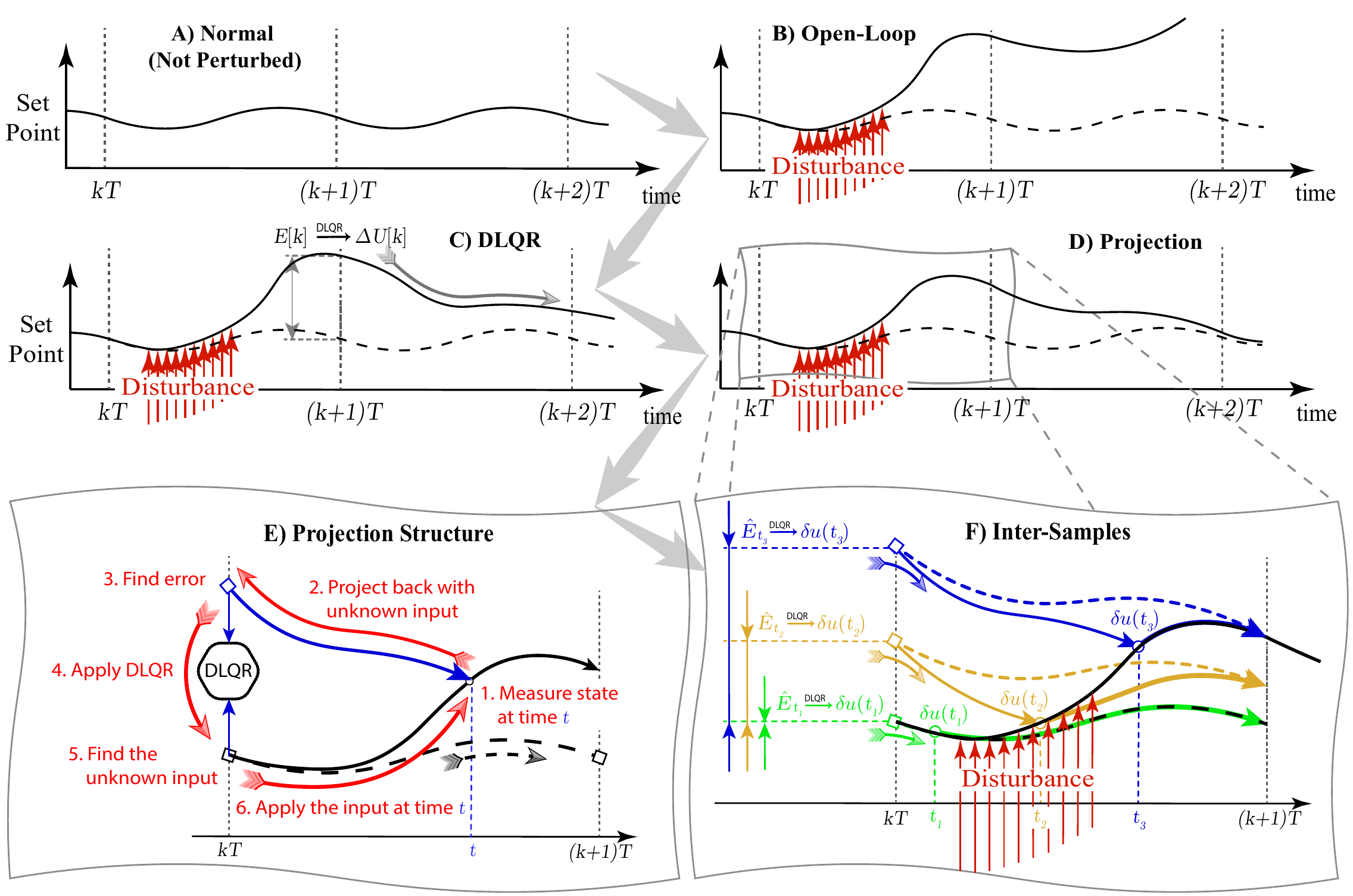}
	\caption{A comprehensive schematic of time-projection idea. Panel A) shows an open-loop system with nominal trajectories. B) This system might not be stable when subject to disturbances. C) A DLQR controller designed for events $kT$ can stabilize this system but provides corrections only at times $kT$. Such a delayed reaction might produce large state deviations. D) The projection controller reduces this deviation by reacting to the disturbance immediately. Such reaction is done through an online control scheme where the control rate is faster. E) The projection controller is based on projecting inter-sample errors back in time to the discrete events. The DLQR controller can find appropriate corrective inputs for the system based on the projected errors. These inputs are then applied to the system in the inter-sample time instance. F) The projection controller is very much similar to the DLQR controller. Without disturbance at $t_1$, it produces no adjustment. During the disturbance, e.g., at $t_2$, this controller keeps updating the input. When the disturbance is finished, e.g., at $t_3$ the inputs remain constant until the next event. } 
	\label{fig::projecting_idea}
\end{figure*}
To present the idea behind time-projection in a simple way, we consider a linear time-invariant system in which an error should be regulated to zero at certain sampling times. Define a state vector $x(t) \in \mathbb{R}^{N}$ and a control vector $u(t) \in \mathbb{R}^{M}$. The system can be described by:
\begin{eqnarray}
\dot{x}(t) = a x(t) + b u(t)
\label{eqn::simple_system}
\end{eqnarray}
where $a \in \mathbb{R}^{N \times N}$ and $b \in \mathbb{R}^{N \times M}$ are constant matrices. The closed-form solution of this system at time $t$ is obtained by:
\begin{eqnarray}
x(t) = e^{at} x(0) + \int_0^t \! e^{a(t-\tau)} b u(\tau) \, \mathrm{d}\tau
\label{eqn::simple_system_solution}
\end{eqnarray}
For simplicity, we consider a constant input here, although this can be extended to linear or quadratic forms without loss of generality. With a constant input, parametrized by the vector $U \in \mathbb{R}^{M}$, the equation (\ref{eqn::simple_system_solution}) takes the form:
\begin{eqnarray}
x(t) = e^{at} x(0) + (e^{at} - I)a^{-1}b \, U
\label{eqn::simple_system_constantU}
\end{eqnarray}
where we assumed $a$ is invertible. If $a$ is singular, a similar expression can be obtained by considering the Jordan form of $a$. We consider a period time $T>0$ at which the behavior of this system could be described discretely: 
\begin{eqnarray}
X[k+1] = A(T) X[k] + B(T) U[k]
\label{eqn::simple_system_discrete}
\end{eqnarray}
where $X[k]=x(0)$, $X[k+1]=x(T)$, $A(T) = e^{aT}$ and $B(T) = (e^{aT} - I)a^{-1}b$. 

\subsection{DLQR controller}

Assume this system has a nominal solution $\bar{X}[k]$ and input $\bar{U}[k]$. Due to linearity, we can define error dynamics as follows:
\begin{eqnarray}
E[k+1] = A E[k] + B \Delta U[k]
\label{eqn::simple_system_error dynamics}
\end{eqnarray}
where $E[k] = X[k] - \bar{X}[k]$ and $\Delta U[k] = U[k] - \bar{U}[k]$. Here we dropped function inputs $(T)$ from $A(T)$ and $B(T)$ for simplicity. Assume this system is controllable and a DLQR controller can be found to minimize the following cost function and constraint:
\begin{eqnarray}
\nonumber &\underset{E[k],\Delta U[k]}{\text{min}} \sum_{k=0}^{\infty} E[k]^TQE[k]+\Delta U[k]^TR\Delta U[k] \\
& s.t. \, \begin{array}{l} E[k+1] = AE[k]+B\Delta U[k]  \end{array}
\, \, \,  k\ge0
\label{eqn::simple_system_lqr}
\end{eqnarray}
where $Q$ and $R$ are cost matrices. The optimal gain matrix calculated from this optimization is called $K \in \mathbb{R}^{M \times N}$, producing a correcting input $\Delta U[k] = -KE[k]$.

Conceptual trajectories plotted in Figure \ref{fig::projecting_idea}.A demonstrate the set point nominal solution $\bar{X}[k]$ and $\bar{U}[k]$ without any disturbance. As shown in Figure \ref{fig::projecting_idea}.B, in presence of disturbances, this open-loop system might be unstable and easily diverge from nominal trajectories. The nature of DLQR controller designed earlier is shown in Figure \ref{fig::projecting_idea}.C. This controller is only active at time instants $kT$ and provides corrective inputs to the system until the next sample $(k+1)T$. In this period, the effect of any disturbance on the system is cumulative, without any correction. Because of an exponential nature in (\ref{eqn::simple_system_solution}), even a small intermittent disturbance might create a substantial error at time $(k+1)T$. This issue might be fixed by increasing the resolution and redesigning the DLQR controller depending on disturbance dynamics. However, we can take advantage of the already designed DLQR controller and increase the control samples easily.

\subsection{Projection controller}

Thanks to closed-form system equations (\ref{eqn::simple_system_constantU}), we can map the state measured at any time $t$ to the samples before and after. Figure \ref{fig::projecting_idea}.E shows projection controller idea between samples $kT$ and $(k+1)T$. In brief, to apply time-projection:
\begin{enumerate}
	\item Measure the current state $x(t)$ at time $t$.
	\item Project $x(t)$ back in time with an unknown $\delta \hat{U}_t[k]$ to find a possible initial state $\hat{X}_t[k]$.
	\begin{eqnarray}
	\nonumber x(t) =&& A(t-kT) \hat{X}_t[k] \\
	+ \, && B(t-kT) \, (\bar{U}[k]+\delta \hat{U}_t[k])
	\label{eqn::simple_system_xtoX+}
	\end{eqnarray}
	Here, hat notation means that these variables are just predicted at time $t$ and they are not actual system variables. The subscript $t$ also indicates dependency on $t$.
	\item Now, given $\hat{X}_t[k]$ and $\bar{X}[k]$, find a projected error $\hat{E}_t[k]$ at the beginning of the phase. 
	\begin{eqnarray}
	\hat{E}_t[k] = \hat{X}_t[k] - \bar{X}[k]
	\label{eqn::simple_system_X-toE}
	\end{eqnarray}
	\item The expertise of DLQR controller can be used here to apply a feedback on $\hat{E}_t[k]$.
	\begin{eqnarray}
	\delta \hat{U}_t[k] = -K \hat{E}_t[k]
	\label{eqn::simple_system_dU}
	\end{eqnarray}
	\item Assuming that the output of DLQR is the same unknown $\delta \hat{U}_t[k]$, one can solve a system of linear equations to find $\delta \hat{U}_t[k]$. Defining $\bar{x}(t) = A(t-kT) \bar{X}[k] + B(t-kT) \bar{U}[k]$ and $e(t) = x(t) -\bar{x}(t)$ we have:
	\begin{eqnarray}
	\begin{bmatrix} A(t-kT) & B(t-kT) \\ K & I \end{bmatrix} \begin{bmatrix}
	\hat{E}_t[k] \\ \delta \hat{U}_t[k] \end{bmatrix} = \begin{bmatrix} e(t) \\ 0
	\end{bmatrix}
	\label{eqn::simple_system_solvedU}
	\end{eqnarray}
	\item The input is then directly transferred to time $t$ without any change and applied to the system:
	\begin{eqnarray}
	\delta u(t) = \delta \hat{U}_t[k]
	\label{eqn::simple_system_dUanswer}
	\end{eqnarray}
\end{enumerate}

This simple procedure can be done for any time instance $t$ between the coarse time-samples $kT$ and $(k+1)T$ to update the control input. The projection loop involves solving a linear system of equations whose dimensions depend on the number of control inputs and states in the system. Figure \ref{fig::projecting_idea}.F demonstrates three different inter-sample times $t_1$, $t_2$ and $t_3$ before, in the middle and after the disturbance respectively. When following nominal trajectories, the projection controller produces zero adjustments $\delta u(t)$ indeed. During the disturbance, the controller keeps updating $\delta u(t)$, and after the disturbance, it produces the same adjustment every time until the end of the phase. The resulting conceptual trajectories are shown in Figure \ref{fig::projecting_idea}.D which naturally outperform the DLQR (Figure \ref{fig::projecting_idea}.C) due to a higher control rate.

A detailed demonstration of projection control is provided in Appendix C for a simple system. We discuss continuous and discrete control designs for this simple system and provide a numerical example. DLQR feedback gains in our case always fall within the range of allowable gains which stabilize the closed-loop discrete system. Projection control, however, results in a time-variant continuous feedback which is absent in continuous and discrete LQR paradigms. This property slightly shrinks the allowable range to avoid infinite continuous feedback gains. We precisely discuss this issue in Appendix C to provide further insights. For our walking model and the choice of dimension-less DLQR cost design, this property is always satisfied through simulations. Within the allowable range of feedback gains, a stability proof is straight-forward for the simple system as provided in Appendix C. We plan to extend this proof to higher dimensions in a follow-up paper with more detailed mathematics discussions. 

\section{3LP Model}

\begin{figure}[]
	\centering
	\includegraphics[trim = 0mm 0mm 0mm 0mm, clip, width=0.25\textwidth]{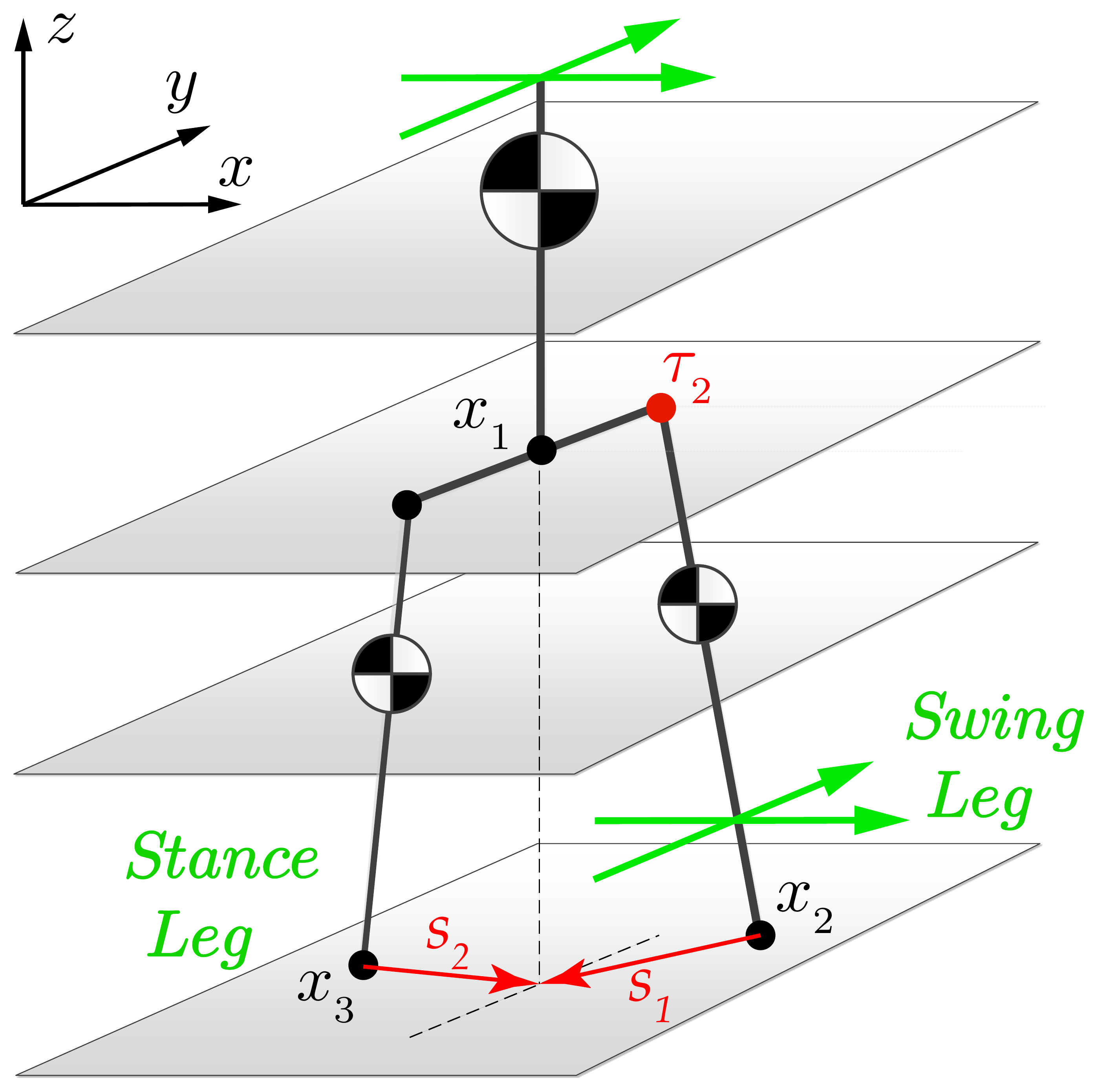}
	\caption{The 3LP model composed of three limbs to represent the torso and the two legs. The three masses, pelvis and the two feet all remain in constant height planes, assuming ideal prismatic actuators in the legs. Horizontal positions of $x_1$, $x_2$ and $x_3$ and their derivatives determine the state of 3LP at each instance of time. In this paper, we only consider swing hip torques as inputs to the system. These torques are applied to the swing leg in frontal and sagittal planes. The horizontal vectors $s_1$ and $s_2$ are also defined to quantify the symmetry in the gait over consecutive motion phases. }
	\label{fig::3lp}
\end{figure}

A suitable application of the projection controller would be a walking system due to the presence of hybrid phases. For this purpose, we use a linear model called 3LP \cite{faraji20173lp} in simulations to demonstrate the functionality of time-projection and to provide physical intuitions. The 3LP model is a simple extension of LIP model by adding a pendulum to represent the torso and another inverted pendulum to represent the swing leg \cite{faraji20173lp}. 3LP has three masses, one located in the middle of each pendulum. In each limb, there are two prismatic actuators like the LIP model with ideal controllers to keep the mass and the pelvis both at constant height levels. There is a massless pelvis of a certain width connecting the two legs with the torso rigidly connected in the middle. Model geometries in 3LP are taken from human anatomical data \cite{de1996adjustments, faraji20173lp}. Figure \ref{fig::3lp} demonstrates the 3LP model in a 3D space. This model can simulate periodic frontal and sagittal motions of walking as well as transient conditions by changing inputs, i.e., swing hip and stance ankle torques. In 3LP, all stance leg hip torques are calculated to avoid torso pitch and roll, assuming ideal controllers in the hip joint. The rest of internal forces and torques are calculated as a function of input torques to realize model assumptions (constant heights and fixed torso). Motion equations are then derived from solving Newtonian mechanics for this system. 

\subsection{Mechanics}

Mechanical equations are written for each pendulum of 3LP in the Cartesian space. We approximated limbs with thin rods of zero inertia along the rod axis, but a non-zero inertia along the two other axis, i.e. in frontal and sagittal directions. We skip the details here and refer to the original paper \cite{faraji20173lp} for further information. In brief, following notations of Figure \ref{fig::3lp}, we can define:
\begin{eqnarray}
	x(t) = \begin{bmatrix} x_2(t) \\ x_1(t) \\ x_3(t) \end{bmatrix},\ \ u(t) = \begin{bmatrix} \tau_{2,y}(t) \\ \tau_{2,x}(t) \end{bmatrix} = u_{c} + t\ u_{r}
\end{eqnarray}
where $x_i \in \mathbb{R}^2$ for $i=1,2,3$ represent horizontal positions and $u_c,u_r \in \mathbb{R}^2$ are defined to parametrize swing hip torques with piecewise linear profiles. With these vector variables, one can find system equations in the single support phase by combining equations and resolving internal forces and torques:
\begin{eqnarray}
	\frac{d^2}{dt^2}{x}(t) = C_x \ x(t) + C_u \ u(t) + C_d \ d
	\label{eqn::continuous_3lp}
\end{eqnarray}
which is a simplified version of original equations used in \cite{faraji20173lp}. The variable $d=\pm1$ indicates right or left support which is used to describe the effect of pelvis width. In this paper, we do not include the double support phase and ankle torques for simplicity, although the idea of time-projection can be easily extended to those conditions as well. Note that since $x_3(t)=const$ represents the stance foot position, the corresponding accelerations are automatically set to zero in (\ref{eqn::continuous_3lp}). 

\subsection{Closed-Form Solutions}

The continuous-time linear DAEs of (\ref{eqn::continuous_3lp}) can be solved in closed-form as:
\begin{eqnarray}
	q(t) = A(t) q(0) + B(t) u + C(t) d
	\label{eqn::discrete_3lp}
\end{eqnarray}
where $q(t) \in \mathbb{R}^{12}$ and $u \in \mathbb{R}^4$ are:
\begin{eqnarray}
	q(t) = \begin{bmatrix} x(t) \\ \dot{x}(t) \end{bmatrix} ,\ u = \begin{bmatrix} u_{c} \\ u_{r} \end{bmatrix}
\end{eqnarray}
Note that equations (\ref{eqn::continuous_3lp}) and (\ref{eqn::discrete_3lp}) are defined for stance and swing legs, not for left and right legs. At every phase change, a matrix $S \in \mathbb{R}^{12 \times 12}$ is multiplied to $q(t)$ to exchange the two legs:
\begin{eqnarray}
	S =& \begin{bmatrix} S_x & \cdot \\ \cdot & S_{x} \end{bmatrix} \\
	S_x =& \begin{bmatrix} 	\cdot & \cdot & I_{2\times2} \\
							\cdot & I_{2\times2} & \cdot \\
							I_{2\times2} & \cdot & \cdot 
		  \end{bmatrix}
\end{eqnarray}

\subsection{Periodic Solutions}

In the beginning and at the end of each phase, all foot velocities are zero, i.e. $\dot{x}_i(0)=0$ and $\dot{x}_i(T)=0$ for $i=2,3$. Here, $T$ represents the phase duration. To find periodic solutions, we consider symmetric gaits where feet positions are symmetric with respect to the pelvis in the beginning of each phase. To realize this symmetry, we define a matrix $M$ as:
\begin{eqnarray}
	M =& \begin{bmatrix} M_x & \cdot \\ \cdot & M_x	\end{bmatrix} \\
	M_x =& \begin{bmatrix}   -I_{2\times2} & I_{2\times2} & \cdot \\
							 \cdot & I_{2\times2} &-I_{2\times2} 
	\end{bmatrix} 
	\label{eqn::symmetry}
\end{eqnarray}
which extracts symmetry vectors $s_1$ and $s_2$ and their derivatives shown in Figure \ref{fig::3lp}. A similar matrix $N$ sets initial velocities to zero:
\begin{eqnarray}
	N = \begin{bmatrix} 
						 \cdot & \cdot & \cdot & I_{2\times2} & \cdot & \cdot  \\
						 \cdot & \cdot & \cdot & \cdot & \cdot & I_{2\times2} 
	\end{bmatrix}
\end{eqnarray}
The symmetry concept in the lateral plane requires a negative sign. Therefore, a matrix $O$ is also needed:
\begin{eqnarray}
	O = diag(\ [1,-1, 1,-1, 1,-1, 1, -1]\ )
\end{eqnarray}
With these definitions, any periodic initial state vector $\bar{Q}[k]$, inputs $\bar{U}[k]$ and direction $\bar{D}$ should satisfy the following equation as a function of phase period $T$:
\begin{eqnarray}
	\nonumber M \bar{Q}[k] &=& OMS(A(T)\bar{Q}[k]+B(T)\bar{U}[k]+C(T)\bar{D}) \\
	N \bar{Q}[k] &=& 0
	\label{eqn::3lp_periodic}
\end{eqnarray}
which enforces symmetry at the beginning of two consecutive phases and sets initial velocities in $\bar{Q}[k]$ to zero. Note that the matrix $A(T)$, by design, does not move the stance leg position $x_3(t)$. Note also that $\bar{D}=\pm 1$ and considering dimensions of $\bar{Q}[k]$ and $\bar{U}[k]$, there are four degrees of freedom left in equations (\ref{eqn::3lp_periodic}). The null-space of (\ref{eqn::3lp_periodic}), therefore, defines a class of walking gaits with different speeds and hip-torque profiles. In the original 3LP paper \cite{faraji20173lp}, we enforce a certain gait velocity and minimize all hip torques to solve for redundancy. We use the same concept here, although one could also enforce a certain step-width, CoP profile with ankle torques and etc. 

\subsection{Linearization}

Since the system is essentially linear, this section refers to derivation of error evolution equations which are the same for any walking gait satisfying (\ref{eqn::3lp_periodic}). Consider a vector $x(t) \in \mathbb{R}^8$ which extracts symmetry vectors $s_1$ and $s_2$ and their derivatives (shown in Figure \ref{fig::3lp}) from the full state vector $q(t)$:
\begin{eqnarray}
	x(t) = \begin{bmatrix} s_1 \\ s_2 \\ \dot{s}_1 \\ \dot{s}_2 \end{bmatrix} = M q(t)
	\label{eqn::reduced}
\end{eqnarray}
This vector is in fact the reduced state vector with respect to the stance foot position. Now, consider another matrix $\hat{M}$ similar to $M$ which distributes a reduced state vector $x(t)$ over the full vector $q(t)$:
\begin{eqnarray}
	\hat{M} =& \begin{bmatrix} \hat{M}_x & \cdot \\ \cdot & \hat{M}_x	\end{bmatrix} \\
	\hat{M}_x =& \begin{bmatrix}   -I_{2\times2} & I_{2\times2} \\
								   \cdot & I_{2\times2} \\ 
								   \cdot & \cdot \\ 
	\end{bmatrix} 
	\label{eqn::symmetry_reverse}
\end{eqnarray}
It is obvious that $M\hat{M} = I_{8\times8}$, since $\hat{M}$ expands the reduced vector and $M$ shrinks it again. Due to linearity and thanks to closed-form solutions, we can easily derive equations that describe evolution of errors or deviations from nominal periodic solutions. Consider a reduced initial error vector $E[k] \in \mathbb{R}^8$ which is added to a nominal periodic solution $\bar{Q}[k]$ by $Q[k] = \bar{Q}[k]+\hat{M}E[k]$. The deviated vector $Q[k]$ evolves in time according to (\ref{eqn::discrete_3lp}):
\begin{eqnarray}
Q[k+1] =& A(T) (\bar{Q}[k]+\hat{M}E[k]) \\
+&B(t) (\bar{U}[k]+\Delta U[k]) + C(T)\bar{D}
\label{eqn::error1}
\end{eqnarray}
where an additional control input $\Delta U[k]$ is also applied to the system. The nominal gait $\bar{Q}[k]$ evolves by:
\begin{eqnarray}
\bar{Q}[k+1] = A(T) \bar{Q}[k] + B(T) \bar{U}[k] + C(T)\bar{D}
\label{eqn::error2}
\end{eqnarray}
The error at time $T$ can be extracted from the transferred $Q[k+1]$ vector by:
\begin{eqnarray}
E[k+1] = OMS(Q[k+1] - \bar{Q}[k+1])
\label{eqn::error3}
\end{eqnarray}
Equations (\ref{eqn::error1}), (\ref{eqn::error2}) and (\ref{eqn::error3}) altogether define error evolution dynamics:
\begin{eqnarray}
\nonumber E[k+1] =& \hat{A}(T) E[k] + \hat{B}(T) \Delta U[k] \\
\nonumber \hat{C} E[k+1] =& 0 \\
s.t. &\left\{
\begin{array}{ccc}
\hat{A}(T) &=& O M S A(T) \hat{M} \\
\hat{B}(T) &=& O M S B(T) \\
\hat{C} &=& \begin{bmatrix} 0_{2\times4} & -I_{2\times2} & I_{2\times2} \end{bmatrix}
\end{array}
\right.
\label{eqn::error_dynamics}
\end{eqnarray}
Note that the terminal swing foot velocity should be zero with respect to the stance foot (i.e. $\dot{s}_2(T)-\dot{s}_1(T)=0$). Therefore, a constraint matrix $\hat{C}$ should be applied to $E[k+1]$ all the time. Equations (\ref{eqn::error_dynamics}) are used directly in formulations of Appendices A and B for DLQR and projection control.  


\section{Results}

In this section, we demonstrate walking scenarios which show advantages of using the projection controller over the DLQR controller. These scenarios include simulating intermittent pushes and speed tracking where periodic system trajectories change over time. All simulations are done with a human-sized 3LP model in Matlab which has a mass of $m=70kg$ and a height of $h=1.7m$. Our choice of DLQR cost coefficients are dimensionless diagonal cost matrices $Q = I_{8\times8}$ and $R = (mg)^{-2} I_{4\times4}$ for states and inputs, where $g$ is gravity. All results and qualitative conclusions of this section are in fact valid for different 3LP models matching humanoid robots like Atlas with $m=150kg$ and $h=1.88m$ \cite{Feng2014optimization}, Coman with $m=30kg$ and $h=1m$ \cite{coman} and Walk-Man with $m=120kg$ and $h=1.85m$ \cite{tsagarakis2017walk}. We present simulation results of a human-sized 3LP only for simplicity.

\subsection{Intermittent pushes}

\begin{figure*}[]
    \centering
    \includegraphics[trim = 0mm 0mm 0mm 0mm, clip, width=\textwidth]{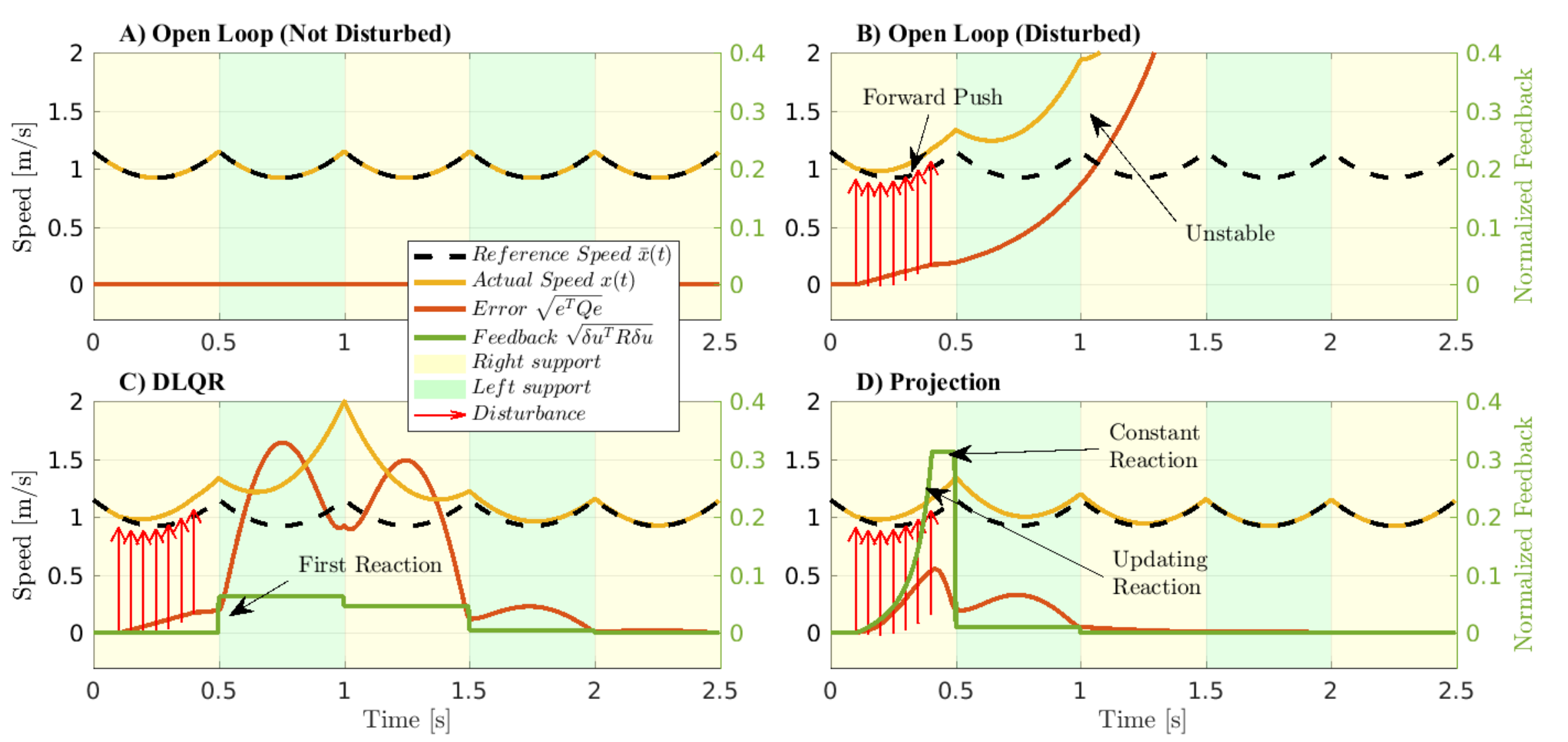}
    \caption{Trajectories of A) open-loop normal, B) open-loop disturbed, C) DLQR and D) projection controllers. Due to falling dynamics and unstable system modes, a moderate disturbance can lead to falling in few steps. The DLQR controller updates inputs only at phase-change times which produces large deviations due to a delayed reaction. The projection controller, however, resolves this issue by a fast online reaction. Note that the projection controller is only using the knowledge of the DLQR controller and uses projection to translate continuous online errors to discrete errors. The expertise of DLQR controller on discrete errors is then used to produce online corrections.} 
    \label{fig::projecting_walking}
\end{figure*}

Consider a gait with a frequency of $f = 2 \ steps/s$ and a speed of $v = 1\ m/s$. The reduced system state (\ref{eqn::reduced}) has eight dimensions, though we only use the sagittal pelvis velocity as a periodic signal for demonstrations. The open-loop velocities are shown in Figure \ref{fig::projecting_walking}.A where the reference nominal trajectory $\bar{x}(t)$ matches the open-loop system behavior. Now, applying an inter-sample forward push in the first phase of the motion results in an unstable deviation from nominal trajectories which eventually lead to falling as shown in Figure \ref{fig::projecting_walking}.B. This is because of unstable open-loop eigenvalues due to the natural falling dynamics. The DLQR controller can stabilize the system by adjusting the footstep at time $t=1$, shown in Figure \ref{fig::projecting_walking}.C. This adjustment is a result of adjusting swing hip torques at the beginning of the phase (at $t=0.5$), where the accumulated deviation was first detected. As a result, corrective inputs $\Delta U$ are nonzero after the time $t=0.5$ until the system stabilizes completely. 

The projection controller has a different behavior, however. Without deviations, this controller produces no corrective input. Once a deviation exists, if there is no active disturbance applied to the system, the output of projection controller remains the same. However, once an external disturbance is present, the projection controller keeps updating the inputs which can be seen in Figure \ref{fig::projecting_walking}.D. Notice the first phase where the external push starts acting on the system at $t\approx0.1$. At this time, corrective inputs start increasing until the disturbance disappears at $t\approx0.4$. After, the correction produced by the projection controller remains the same until $t=0.5$. In the next phase, the new stance foot location is already adjusted, and thus less effort is needed by the controller. In other words, online adjustment of the swing hip torque in the first phase places the swing foot in a proper location so that the second phase starts in good condition. Such footstep adjustment produces a larger error norm in the first phase compared to the DLQR controller, however, reduces this norm in the next phases considerably. As a result, overall deviations from nominal trajectories are less with the projection controller, thanks to an online updating scheme. 

\begin{figure*}[]
    \centering
    \includegraphics[trim = 0mm 0mm 0mm 0mm, clip, width=1\textwidth]{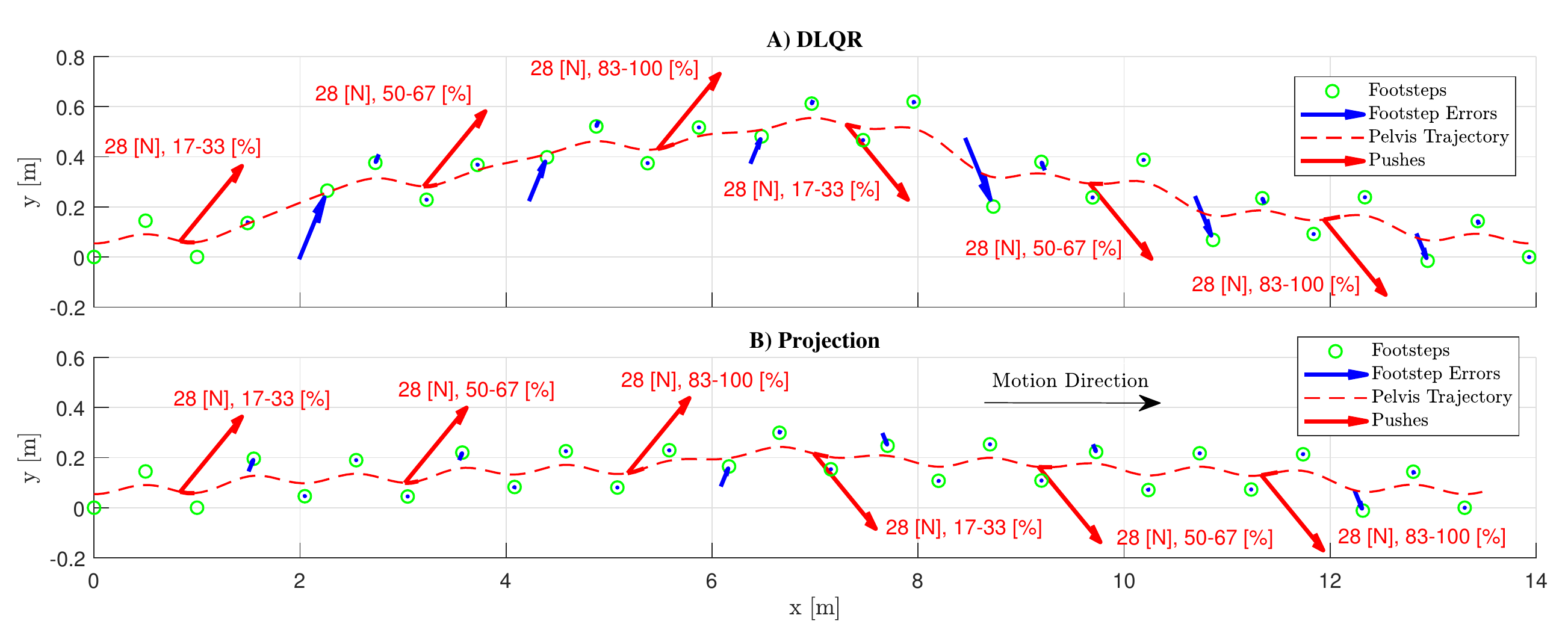}
    \caption{ Performance of A) the DLQR and B) the projection controllers in rejecting intermittent external pushes. In this scenario, we apply a push with the same magnitude and duration, but at different times during the phase. We also apply the same pushes in the opposite lateral direction when the other leg is in swing phase, and the CoM has a different lateral motion direction. Along with the force magnitudes, we show by percentage, the period in which the push is applied. Like Figure \ref{fig::projecting_walking}, the reaction of DLQR controller is only taking place after the push with a delay. It is also notable that a pushing force at the end of the phase has a relatively small impact. However, the same push at the beginning and in the middle of the phase can result in larger corrective steps. The lateral direction of the push does not influence the size of corrective steps, since the error equations (\ref{eqn::error_dynamics}) are linear and symmetric in both directions. Likewise, a backward push results in the same footstep adjustment magnitude. The corresponding videos could be found in Multimedia Extension 1.} 
    \label{fig::partial_pushes}
\end{figure*}

A better understanding of time-projection behavior for the walking application can be obtained through a direct plot of actual footstep locations. For this purpose, we reduced the speed to $v=0.5\ m/s$ to plot more footsteps in a single figure and simulated pushes of same magnitude and duration but applied at different times during the phase. The resulting footstep plans are demonstrated in Figure \ref{fig::partial_pushes} for both DLQR and projection controllers. As expected, late pushes have less impact while early pushes can cause large corrective steps with the DLQR controller, shown in Figure \ref{fig::partial_pushes}.A. This can be confirmed by the exponential effect of sensory or model errors in the forward integration of our equations \cite{bhounsule2015discrete}. With the DLQR controller, the reaction is only taken in phase-transition events which can introduce delays if the push is applied much earlier, at the beginning of the phase. Although the controller can still stabilize the system, large input torques or step lengths are not desired in practice. The projection controller adjusts the first footstep location online during the phase when the push is being applied. The swing leg has, in fact, a negligible influence on the system and thus, this adjustment does not provide immediate correction. However, since the swing foot becomes the next stance foot which has a strong coupling to system variables, an early adjustment of its position can considerably stabilize the system in the next phase.

\subsection{Speed Modulation}

\begin{figure}[]
    \centering
    \includegraphics[trim = 0mm 0mm 0mm 0mm, clip, width=0.5\textwidth]{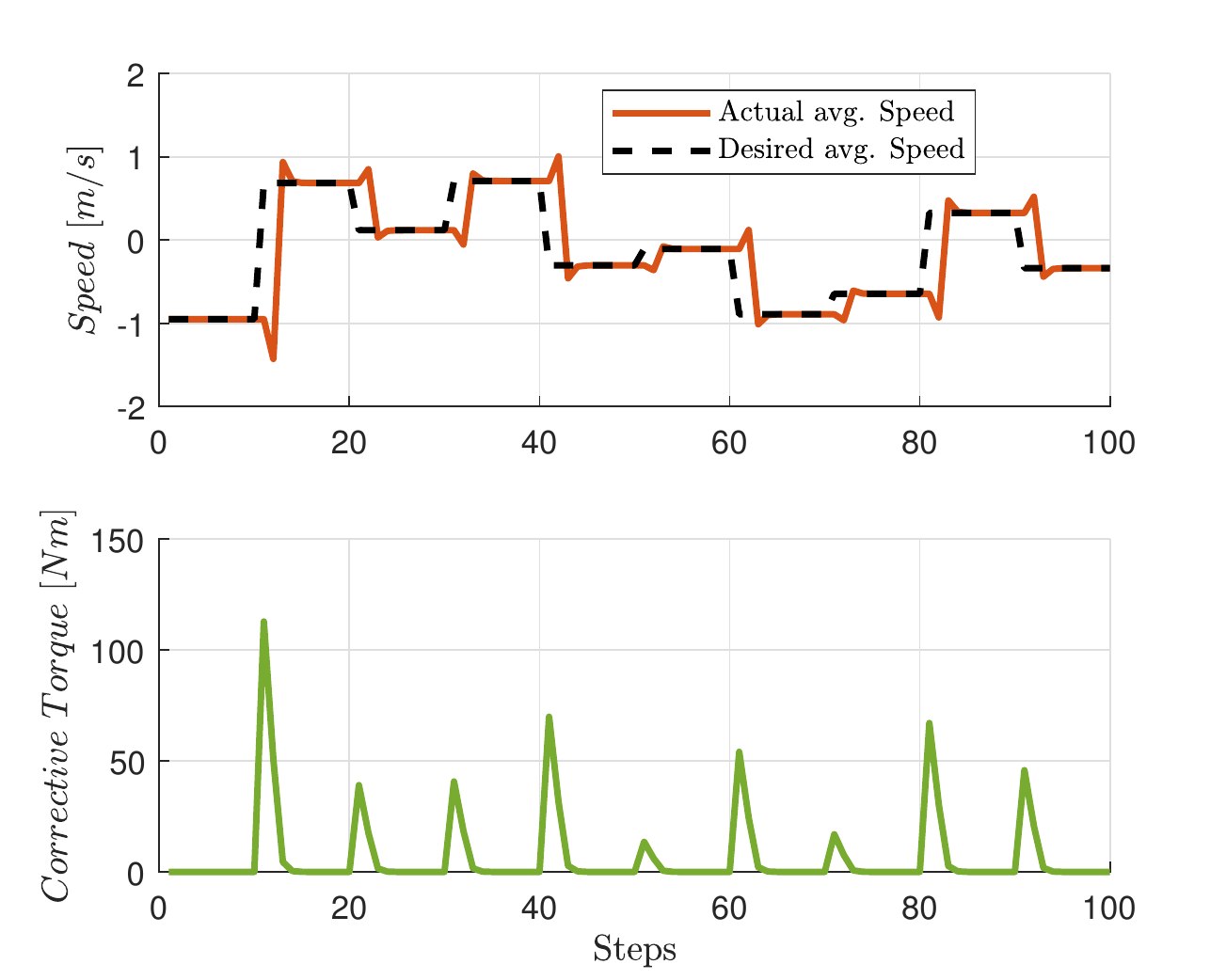}
    \caption{Performance of projection controller in transitioning between different speeds. The original gait is simply scaled to modulate the speed within the null-space provided by equations (\ref{eqn::3lp_periodic}). Since the instantaneous pelvis velocity has variations, we reported the average speed at each discrete event by dividing feet distance with stride time. On the top graph, speed tracking performance is shown for the desired profile of speeds which changes randomly at every ten steps. The bottom graph is showing the root-mean-square of controller outputs with respect to the desired gait at each step. One can see that the controller performs the transition in almost two steps. Since we change desired speeds only at phase-change moments, the DLQR controller would perform exactly the same as the time-projecting controller. An example movie of this scenario could be found in Multimedia Extension 1.}  
    \label{fig::tracking}
\end{figure}

Remember that equations (\ref{eqn::3lp_periodic}) provided a class of gaits with different speeds, but a fixed frequency. In this test scenario, we consider changing the reference gait occasionally and observe the tracking performance of the projection controller. We apply these changes only at phase-transition moments, i.e., at every ten steps to make sure the controller has enough time to converge. In Figure \ref{fig::tracking}, resulting trajectories are demonstrated. In fact, since there is no inter-sample disturbance, the projection controller performs similarly to the DLQR controller. Once a new desired speed is commanded at the beginning of a phase, the reference gait $\bar{Q}[k]$ changes and thus, the controller adjusts the swing hip torques. This adjustment leads to modifying the footstep location at most in the following next two steps. Therefore, our controller tracks new velocities in only two steps. Likewise, in Figure \ref{fig::partial_pushes}, disturbances are captured at most in two steps following the analysis provided in \cite{zaytsev2015two}.   

\section{Analysis}

As motivated earlier in Figure \ref{fig::projecting-vs-mpc}, the goal of projection controller is to achieve online control performance based on a DLQR controller. The projection controller is comparable to the MPC controller we used in \cite{faraji2014robust} in the sense that it can react to inter-sample disturbances immediately whereas the DLQR controller has to wait until the end of the phase. The advantage of projection controller to MPC is relieving the need for online numerical optimizations while considering an infinite horizon. In practice, however, MPC can handle inequality constraints which are absent in the projection controller. In this regard, we are interested to know how well our projection controller can cover the set of viable states. In this section, we provide this analysis together with a quantification of open-loop and closed-loop eigenvalues as well as the sensitivity to the timing of external disturbances. 

\subsection{Eigen Values}

\begin{figure}[]
    \centering
    \includegraphics[trim = 0mm 0mm 0mm 0mm, clip, width=0.5\textwidth]{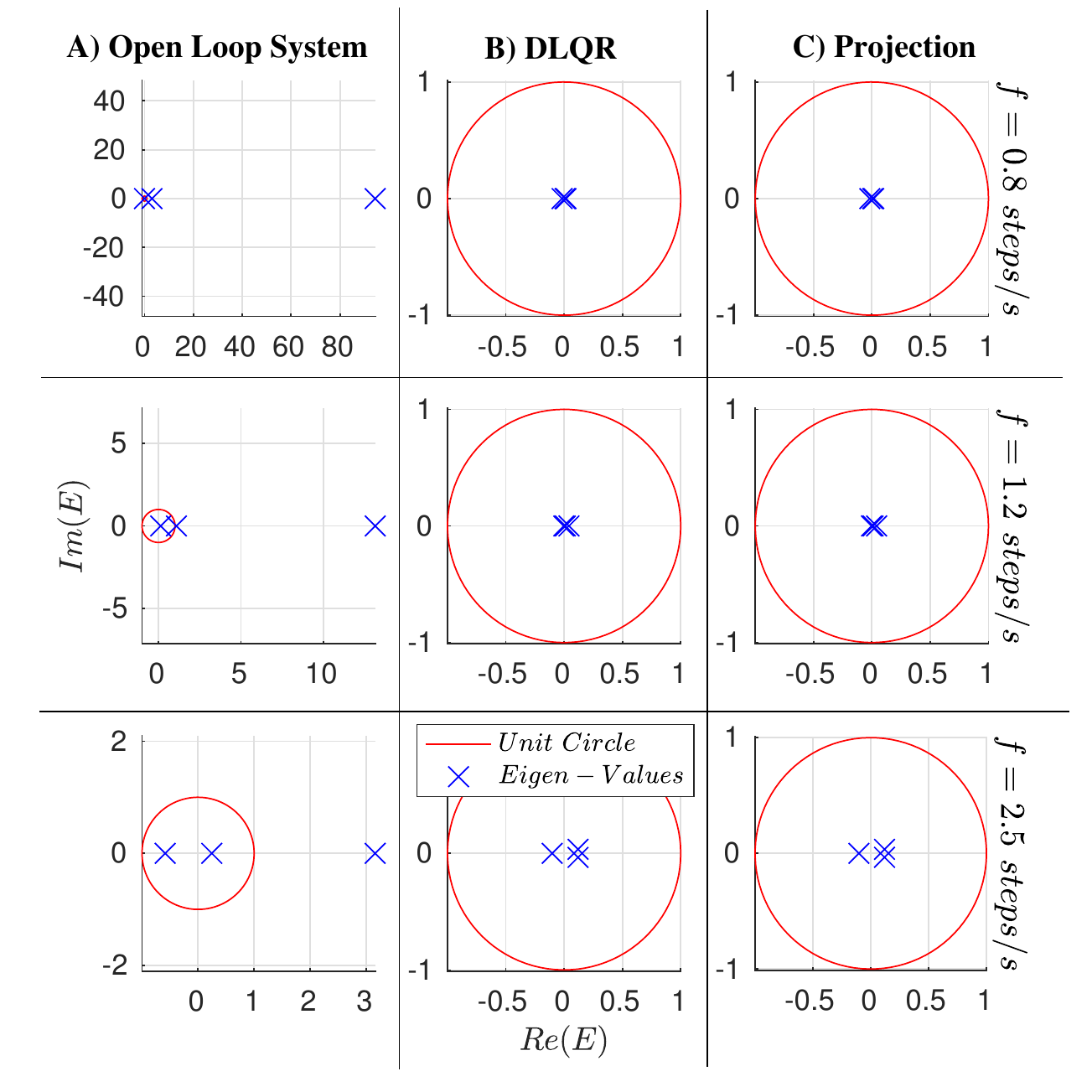}
    \caption{Eigen-values of the walking system in the sagittal direction of $s_1$, $s_2$ and $\dot{s}_2-\dot{s}_1$, calculated over one step. Due to the similarity of lateral and sagittal dynamic equations, lateral eigen-values are duplicates of the plotted sagittal eigen-values here. A) The open-loop system has a large unstable mode depending on the walking frequency. The close-loop systems (with B) DLQR and C) projecting controllers) both stabilize the unstable system through foot-stepping strategy. Note that the DLQR and projecting controllers perform the same in absence of inter-sample disturbances.} 
    \label{fig::eigen}
\end{figure}

Since the linearized model or error dynamics of (\ref{eqn::error_dynamics}) is similar in sagittal and frontal directions, we only consider the sagittal direction and reduce dimensions of $x(t)$ in (\ref{eqn::reduced}) further to four variables only. Also, since the foot velocity is forced to zero in phase-transition moments, this dimension is further reduced to three. Therefore, we only consider evolution of errors in three dimensions (i.e. sagittal directions of $s_1$, $s_2$ and $\dot{s}_2-\dot{s}_1$, shown in Figure \ref{fig::3lp}). To find eigenvalues, we added errors to these three quantities in a phase-transition moment and measured them after one step in open and closed-loop systems. Eigen-values of the resulting Poincar\'e map are shown in Figure \ref{fig::eigen} for the open-loop system, DLQR, and projecting controllers. Due to linearity, such analysis is independent of gait velocity but dependent on the model properties and the gait frequency. Therefore, these plots are repeated for different walking frequencies in a typical range observed in human gaits \cite{bertram2005constrained}. It is observed that both the DLQR and projection controllers can stabilize the unstable open-loop system. Since both perform the same in the absence of inter-sample disturbances, their closed-loop eigenvalues are always similar. 

The walking system considered in this paper is interesting regarding eigenvalues. Because of falling dynamics, this system has a large unstable mode which highly depends on frequency. As observed in Figure \ref{fig::eigen}, the open-loop system at $f=0.8\ steps/s$ has a very large eigenvalue, indicating that the system deviates more if given a longer phase time. In slow walking speeds, humans use a relatively slow frequency determined by a trade-off between falling and swing costs \cite{bertram2005constrained}. Based on our eigenvalue analysis, the third stabilization strategy of adjusting footsteps might not have enough control authority in slow frequencies. Therefore, the first continuous control strategy of modulating the CoP might be more effective in these walking conditions. The second eigenvalue in the open-loop system is small and stable while the third one might be stable or not, depending on the frequency. 

\subsection{Push Recovery Strength}

\begin{figure*}[]
    \centering
    \includegraphics[trim = 0mm 0mm 0mm 0mm, clip, width=1\textwidth]{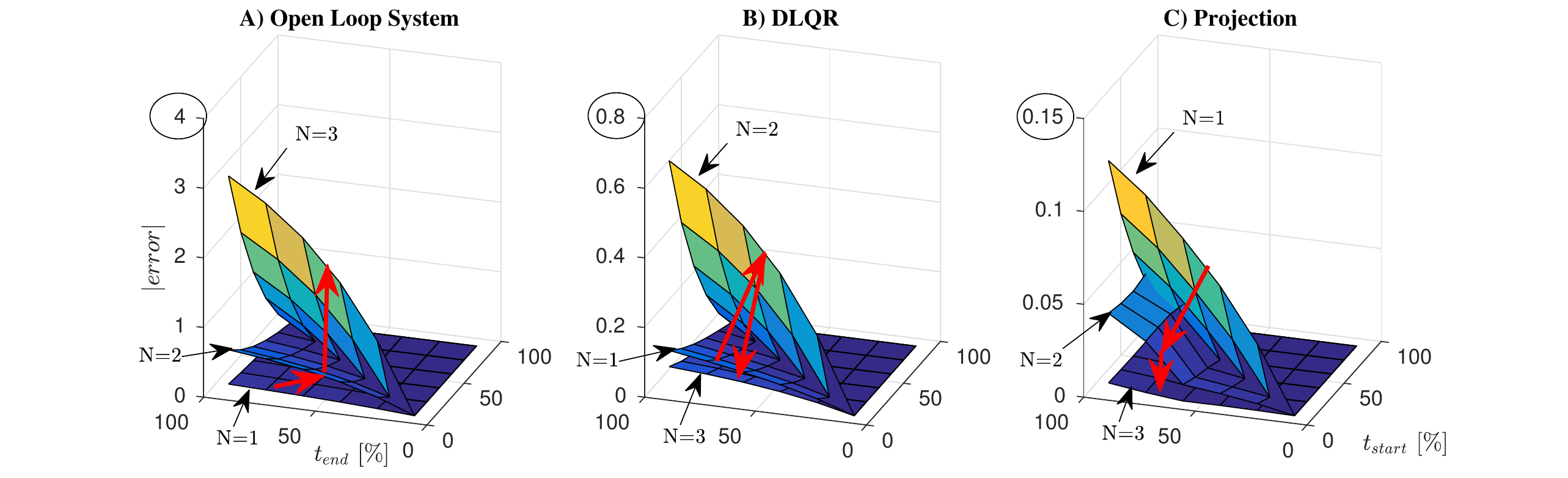}
    \caption{Demonstration of intermittent push recovery performance for the open-loop system, DLQR and projection controllers. In these plots, we show the start and end times of the push application period as a percentage of the phase (like Figure \ref{fig::partial_pushes}). Surfaces show error norms, calculated at touch down events over three consecutive steps. The open-loop system is unstable, the DLQR controller overshoots and the projection controller reacts immediately when the push is being applied. The projection controller is therefore much stronger in rejecting even long-lasting intermittent pushes which span throughout the whole phase.} 
    \label{fig::intermittent}
\end{figure*}

In addition to stability analysis, we would like to characterize how well the projection controller can recover intermittent pushes. In particular, the sensitivity of footstep adjustments with respect to the timing of external pushes is important. To this end, we simulated the open-loop system, DLQR and projection controllers with pushes of the same magnitude, but variable timings and durations like Figure \ref{fig::partial_pushes}. The results are shown in Figure \ref{fig::intermittent} over three consecutive steps. We applied the push between certain times of the phase, shown by percentage on the two horizontal axes. Error surfaces demonstrate the norm of error with respect to the nominal gait, calculated at touch down events and plotted along the vertical axis. 

Figure \ref{fig::intermittent} is demonstrating that a push of same magnitude and duration might have a more severe effect on the system if applied earlier in the phase. While the open-loop system is unstable, the other two closed-loop systems recover the push successfully, but with certain dynamics. At the end of the first step, the DLQR controller produces an error similar to the open-loop system. However, due to a delayed reaction, it overshoots in the second step. The projection controller, however, adjusts the footstep online during the first step to avoid the overshoot. Figure \ref{fig::intermittent} also indicates that longer pushes have more severe effects, especially if they start earlier in the phase. If a push continues until the end of the phase, the projection controller performs slightly worse compared to itself, yet much better than the DLQR controller. 

\subsection{Viable Regions}

Computations in the projection controller involve solving a linear system of equations (\ref{eqn::simple_system_solvedU}) at every fine control time-sample. However, other online controllers like MPC might require more computations due to multiple iterations required to optimize a particular cost function with constraints. The advantage of using MPC over projection is mainly the inclusion of inequality constraints. Without such constraints, the viable region for our linear setup is in fact unlimited. Although the proposed framework ignores these constraints, we would like to characterize the valid set of states in which our controller produces feasible actuator inputs and leads to feasible states, within limitations of the actual hardware.

In this work, we considered constraints on hip torque limits and footstep distances in an adult-size 3LP model. Torque limits are physical limitations, and footstep distances are artificially introduced to prevent violation of constant height assumptions on the real hardware. The torque limits are represented by simple boundaries ($\pm 80 [Nm]$) while next footstep locations should lie on a diamond region centered at the stance foot location (with equal diameters of $1.7m$, compared to a leg length of $0.9m$ in the model). The latter constraint could be more complex like using a circle around the stance foot, but here we used diamonds to preserve linearity of constraints and our analysis. Considering realistic regions indeed require non-convex constraints, especially when foot cross-over and self-collisions are considered.

Due to the coupling of frontal and sagittal motions when adding diamond regions, we cannot split the reduced 6-dimensional discrete error system into 3-dimensional subsystems anymore. Therefore, it is not possible to completely visualize viable regions. Additionally, these regions are now depending on the forward velocity as well. In other words, over faster speeds and thus longer stride lengths, possible foot-adjustments are more limited compared to slower speeds. Therefore, we need to consider the effect of stepping frequency, speeds and dimensionality altogether. With the inequality constraints mentioned earlier, the goal would be to find the maximum set of viable states for each controller \cite{zaytsev2015two} as well as the maximum set for all possible controllers.

Convex polyhedrons of viable regions are calculated for six consecutive steps which seem enough to give an approximate of their complex shape. We also divided each phase into five shorter sub-phases where time-projection and arbitrary input profiles (for maximum viable set) can provide continuous input modifications. The maximum viable set includes all states which are capturable in 6 steps, i.e., with feasible arbitrary inputs and states over these six steps. We observed that increasing steps and resolution exponentially increases the dimensionality and polyhedron computation times while having almost no effect on precision. Since the resulting polyhedrons typically had thousands of vertices, we used a ray-casting method and a linear programmer to search along ray directions and find intersection points with the boundaries. For 2D visualizations, we used a resolution of $100$ rays per $360^o$.

\begin{figure}[]
    \centering
    \includegraphics[trim = 10mm 10mm 5mm 5mm, clip, width=0.5\textwidth]{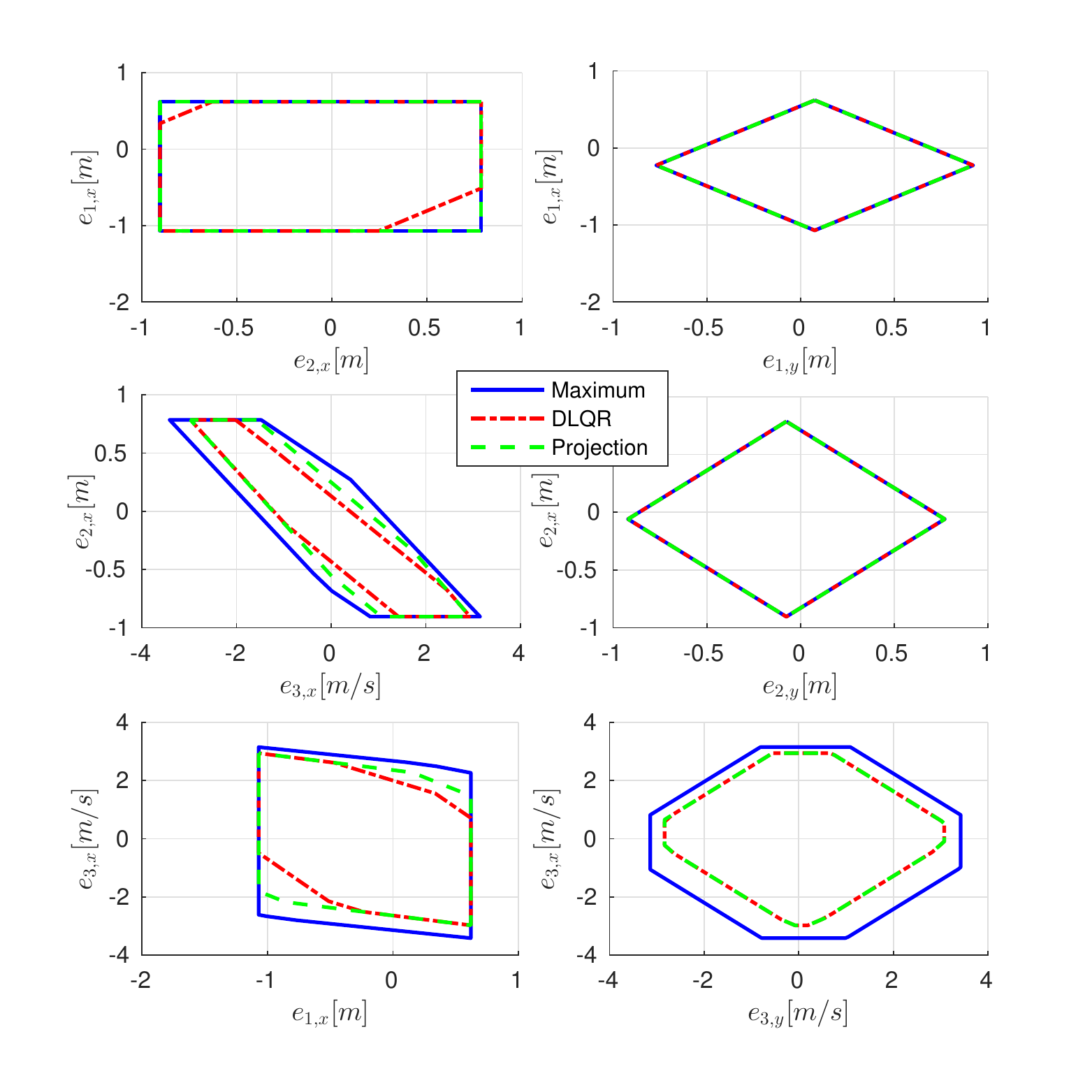}
    \caption{Different spatial projections of 6-dimensional viable regions calculated for the DLQR and projection controllers together with the maximum viable region. The reference gait has a velocity of $v=0.5\ m/s$ and a frequency of  $f=3\ steps/s$. Due to an online update scheme, the projection controller always has bigger viable regions compared to the DLQR controller. In some dimensions, however, the maximum viable region is slightly larger.} 
    \label{fig::sections}
\end{figure} 

Geometric constraints are active at faster speeds while torque limits are active at faster frequencies. It is impossible to view the full polyhedral from all perspectives, but spatial projections on important subspaces are shown in Figure \ref{fig::sections}. We take error vectors $e_1$, $e_2$ and $e_3$ added to the symmetry vectors $s_1$ and $s_2$ and $\dot{s}_2-\dot{s}_1$ (Figure \ref{fig::3lp}) as error measures respectively and plot viable sets with respect to these error dimensions. The diamonds and torque boundary shapes are implicitly observable in some spatial projections while others provide a less intuitive shape. Compared to the DLQR controller, the projection controller has a larger viable set due to an online update scheme. The maximum viable set is slightly larger in some regions though, indicating that there are other types of controllers that can stabilize a slightly bigger set of erroneous states, compared to the projection controller.

\begin{figure}[]
    \centering
    \includegraphics[trim = 5mm 5mm 5mm 5mm, clip, width=0.5\textwidth]{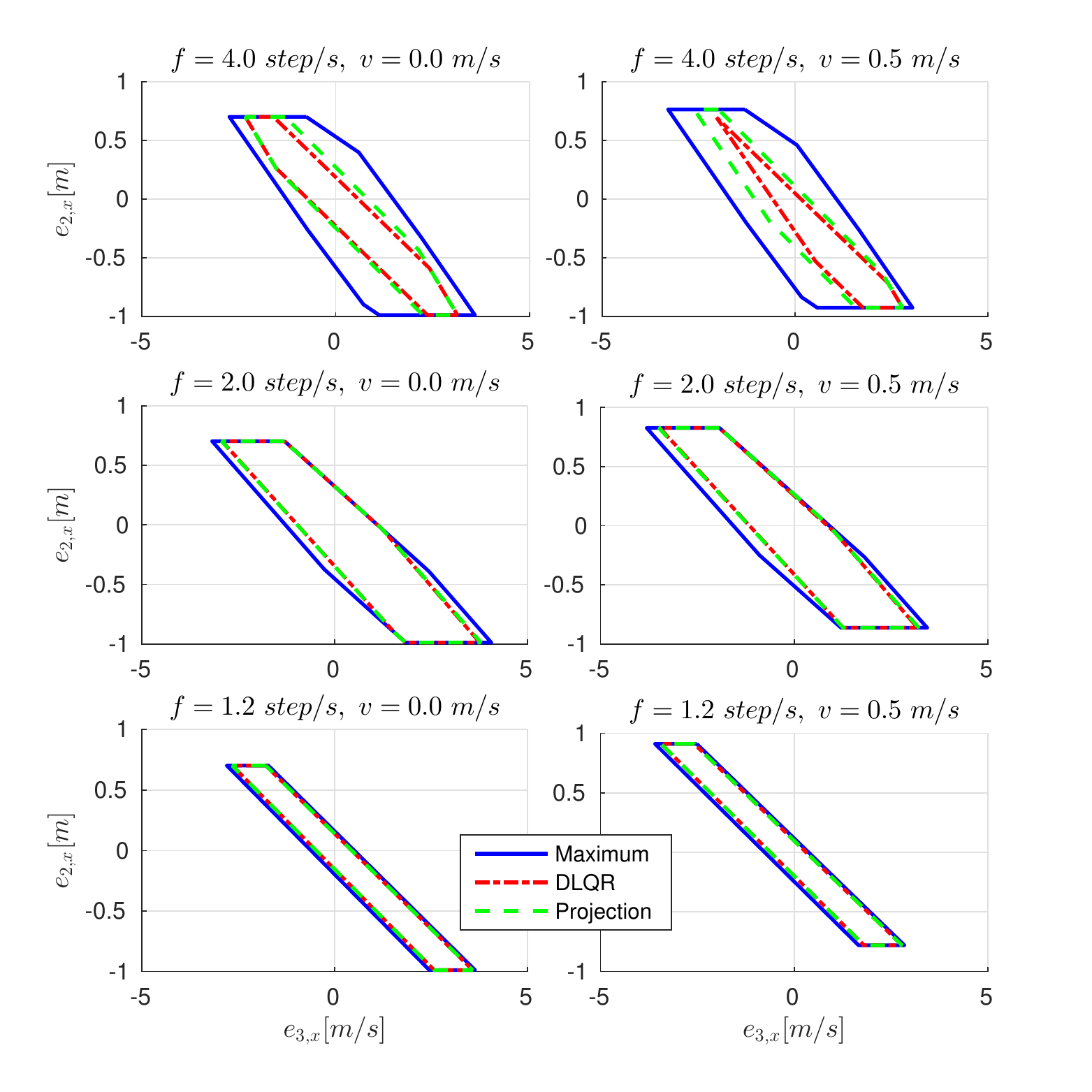}
    \caption{The effect of walking speed and frequency on viable regions. This figure only demonstrates spatial projections of 6-dimensional polyhedrons on one subspace of position-velocity errors. As demonstrated over few choices of velocities and frequencies, the gait velocity mainly shifts the region while faster frequencies can shrink it considerably. } 
    \label{fig::stride_vel}
\end{figure}

Further, we are also interested to see the effect of walking speed and frequency on these regions. Again, it is hard to inspect the full-dimensional polyhedron, though we take a position/velocity spatial projection which is more insightful. Figure \ref{fig::stride_vel} demonstrates viable regions for different choices of walking frequencies and velocities. The velocity mainly shifts the viable region while the frequency can shrink it due to torque limitations. In Figure \ref{fig::stride_vel}, the maximum choice of frequency is unrealistic regarding the human data \cite{bertram2005constrained} just for demonstration purposes. In normal walking conditions, however, the DLQR and time-projection controllers can cover most of the maximum viable region.

\section{Conclusion}

In this paper, motivated by the hybrid nature of walking systems, we formulated a new control framework that can be used for online control. This controller is based on a DLQR controller which works at a rate synchronized to the system's phase-change rate. Such synchrony and a simple derivation of error-dynamic equations which is blind to hybrid phases allow us to use the standard DLQR framework with a fixed model of error evolution. However, in dealing with inter-sample disturbances, this controller cannot be simply used at a higher rate, because of the hybrid phase-change nature. In Appendix C, we show that for normal systems without hybrid phases, increasing the DLQR rate up to a continuous control is always possible. In our hybrid system, however, we introduced the time-projection idea which maps inter-sample errors to discrete errors where they could be easily treated with the DLQR controller. Although time-projection can be used for simple systems as well (refer to Appendix C), increasing DLQR rate and continuous control seem easier for these type of systems. 

\subsection{Physical explanation}
Through extensive simulation and analysis, we showed that time-projection could adjust the footstep location online which naturally outperforms the DLQR controller that has a delayed reaction. This has a benefit in hybrid systems like walking robots since the swing state has little influence on the stability of the system, but as soon as it becomes the next stance foot location, it tightly couples with other system variables. Therefore, investing control effort in adjustment of this state in an online fashion does not improve stability instantaneously (unlike modulating the CoP), but saves a lot of control effort in the next phase of motion. Time-projection helps to translate (projecting) phase-dependent inter-sample errors to phase-free discrete errors where a standard DLQR knows best how to stabilize them. 

\subsection{3LP advantage}
We used 3LP as a walking model to facilitate gait generation and produce more human-like gaits. This was an improvement to the LIP model we used in a previous work \cite{faraji2014robust}. However, the main novelty of this work lies in introduction of the projection controller instead of the MPC controller used in \cite{faraji2014robust}. Footstep adjustment is a known strategy in robotics hopping \cite{raibert1984experiments} and walking \cite{capturability}. Our projection controller generalizes this idea into a DLQR control paradigm where controllers can be designed more systematically. Besides, using 3LP as a physical model helps to find physically more meaningful control laws for footstep adjustment. In 3LP, in fact, footstep adjustment is made through swing hip torques whereas, in other models without a swing leg, this is done directly through adjustment of the final attack angle which translates to footstep location. We believe that the combination of 3LP with time-projection provides a unified control framework in which, in addition to the CoM states, swing location and swing dynamics are also considered. Constraints on modulating the CoP and ankle torques are often used in the literature during trajectory planning \cite{herdt2010online}. However, hip torques are also important to be considered, especially for swing and torso dynamics. 3LP is the first linear template model that can provide information on hip-torques at an abstract level. In \cite{byl2008approximate} for example, since swing dynamics are absent, the hip controller stiffly tracks the desired attack angle only after the mid-stance event. If a hip torque is applied earlier, it can sometimes cause the robot to fall backward. 

\subsection{Comparison to MPC}
The added-value of using time-projection over MPC would be the reduction of computations while considering infinite horizons instead of MPC's receding horizon. However, this framework cannot handle inequality constraints like MPC. Thanks to the analysis of viable regions, we have now a more clear view of other possible controllers. We demonstrated that for normal walking gaits and reasonable constraints imposed by the hardware, time-projection could stabilize the system most of the time. Other controllers can only cover a slightly larger range of disturbed states. If the robot is severely disturbed, inequality constraints of viable regions can simply determine feasibility of a projection control. With such criteria, the algorithm can easily switch to other complicated controllers or even emergency cases. The analysis of viable regions further shows that other controllers can rarely recover if time-projection fails. So, most probably, the robot needs emergency actions. In this regard, MPC controllers cannot do much better than time-projection either, despite considering inequality constraints in their numerical optimization. However, we acknowledge that exploring non-convex inequalities for footstep regions as well as timing adjustment could be possibly handled by nonlinear and non-convex MPC controllers.

\subsection{Different from intermittent control}

As we talk about intermittent disturbances, to avoid confusion, it is worth mentioning that the class of intermittent controllers \cite{gawthrop2007intermittent} are much different from our projection controller. In this work, we propose an architecture based on per time-step measurement and feedback paradigm. Inspired by certain experiments, however, scientists propose a different control architecture for humans \cite{gawthrop2011intermittent}, successfully applied to simple systems too \cite{bhounsule2015discrete}. Instead of traditional observer-predictor-feedback paradigms, intermittent controllers use a kind of feedback which occasionally modifies certain parameters of the low-level feed-forward or feedback controller. Such architecture can better deal with systems in which high-level information is available with lower frequencies or longer delays, like humans. Compared to recent inverse dynamics methods \cite{faraji2014robust}, intermittent control is computationally less demanding of course, but versatility of this approach is yet questionable. Indeed, intermittent control in mixture with Neuro-Muscular model \cite{geyer2010muscle} might provide a reasonable explanation of the control system in humans, especially at slow walking speeds. It is, therefore, a very interesting candidate for the control of 3LP in our future works.

\subsection{Measurement requirements}

In this paper, we assumed the robot state is observable. In other words, the relative pelvis and feet positions and their time derivatives are perfectly measurable. For rigid robots without series elastic elements and precise encoders, this assumption can be realistic. On most of the humanoids, however, a complex Kalman filtering is needed to calculate these states \cite{faraji2015practical}. In this work, the instantaneous error is calculated based on measured positions and velocities while there is no need for a disturbance (external force) observer. However, such information about disturbance can further improve the performance. Knowing that a disturbance exists, the controller can take preventive actions to reduce deviations from nominal trajectories ahead of time.

\subsection{Performance}

Various analyses show that time-projection has similar stabilization and controllability properties with the DLQR controller. However, it performs much better in recovery of intermittent pushes. Overall, advantages of our proposed architecture combined with the 3LP model are:
\begin{description*}
    \item[+] Resistant against intermittent perturbations.
    \item[+] No need for off-line optimization. 
    \item[+] Computationally light compared to MPC.
    \item[+] Online policy refinement.
    \item[+] Slightly expanded viable regions compared to the DLQR.
    \item[+] Covering most of the maximum viable region.
    \item[+] Speed-independent stability analysis. 
    \item[+] Optimal future behavior, thanks to the seeding DLQR.
    \item[+] Generic design for different model sizes, walking speeds and frequencies.
\end{description*}
And disadvantages would be:
\begin{description*}
    \item[-] Fixed timing.
    \item[-] Lack of inequality-constraint support.
\end{description*}

\subsection{Other Applications}

The new projection controller can simply replace our previous MPC controller \cite{faraji2014robust}, bringing many advantages. In future works, we want to use the proposed controller in combination with inverse dynamics and compare it with our previous MPC controllers. We would also like to exploit fast computational properties of the 3LP model to setup nonlinear MPC controllers that can adjust the timing. The inclusion of non-convex constraints for avoiding self-collision is indeed another interesting future extension. Finally, in this paper, we only explored external pushes and not other types of disturbances like uneven terrain. The formulation of 3LP is yet limited to impact-less locomotion which could be extended to handle impacts and height variations. However, the concept of time-projection is yet applicable to other complex models, computationally challenging though if closed-form solutions are not available. Applying time-projection to quadruped locomotion would also be interesting, given that these systems might have more than two distinct phases of motion. This paper is accompanied with a multimedia extension, demonstrating walking motions of speed tracking and intermittent push recovery scenarios. All Matlab codes used in this paper are available online (after being accepted) at \url{http://biorob.epfl.ch/page-99800-en.html}.

\begin{acks}
This work was funded by the WALK-MAN project (European Community's 7th Framework Programme: FP7-ICT 611832).
\end{acks}

\subsection{Appendix A: DLQR for constrained systems}

For a discrete constrained system of:
\begin{eqnarray}
\nonumber &X[k+1] = A X[k] + B U[k]\\
&CX[k+1] = 0
\label{eqn::simple_system_discrete_appA}
\end{eqnarray}
which has a set-point solution $\bar{X}[k]$ and $\bar{U}[k]$, the error system is defined as:
\begin{eqnarray}
\nonumber &E[k+1] = A E[k] + B \Delta U[k] \\
&CE[k+1] = 0
\label{eqn::simple_system_error dynamics_appA}
\end{eqnarray}
where $E[k] = X[k] - \bar{X}[k]$. Here, regardless of the control strategy, the constraint represented by $C$ should always be satisfied. Consider $A \in \mathbb{R}^{N \times N}$ and $B \in \mathbb{R}^{N \times M}$ and $C \in \mathbb{R}^{P \times N}$. The DLQR optimization problem for this system is:
\begin{eqnarray}
\nonumber & \underset{E[k],\Delta U[k]}{\text{min}} \sum_{k=0}^{\infty} E[k]^TQE[k]+\Delta U[k]^TR\Delta U[k]  \\
&s.t. \left\{ \begin{array}{l} E[k+1] = AE[k]+B\Delta U[k] \\ CE[k+1] = 0 \end{array}
\, \, \right. k\ge0
\label{eqn::simple_system_lqr_appA}
\end{eqnarray}
Assume we find a matrix $\tilde{C} \in \mathbb{R}^{(N-P) \times N}$ that forms a complete basis with $C$. In other words, the matrix $S$ defined by:
\begin{eqnarray}
S = \begin{bmatrix} \tilde{C} \\ C \end{bmatrix}
\label{eqn::simple_system_S}
\end{eqnarray}
has a full rank. Now, we define a new variable $Z[k]=SE[k]$ which will produce the following new DLQR problem:
\begin{eqnarray}
\nonumber & \underset{Z[k],\Delta U[k]}{\text{min}} \sum_{k=0}^{\infty} Z[k]^T \tilde{Q} Z[k]+\Delta U[k]^T \tilde{R} \Delta U[k]  \\
&s.t. \left\{ \begin{array}{l} Z[k+1] = \tilde{A} E[k]+ \tilde{B} \Delta U[k] \\ \tilde{C}Z[k+1] = 0 \end{array}
\, \, \right. k\ge0
\label{eqn::simple_system_lqr_Z_optim}
\end{eqnarray}
where:
\begin{eqnarray}
\nonumber \tilde{Q} = S^{-T} Q S^{-1}, \quad \tilde{R} = R \quad \quad \quad\\
\tilde{A} = S A S^{-1}, \quad \tilde{B} = S B, \quad \tilde{C} = C S^{-1}
\label{eqn::simple_system_lqr_Z}
\end{eqnarray}
Note that:
\begin{eqnarray}
Z[k] = SE[k] = \begin{bmatrix} \tilde{C}E[k] \\ CE[k] \end{bmatrix} = \begin{bmatrix} Y[k] \\ 0 \end{bmatrix}
\label{eqn::simple_system_Z_decompose}
\end{eqnarray}
where $Y[k] \in \mathbb{R}^{N-P}$ is reduced to exclude the constraint. Assuming $P<N$, a full rank $C$ and $M>=P$, we can find $P$ independent components of $\Delta U[k]$ and rewrite the constraint in terms of these components. Without loss of generality, assume these components are the last P components of $\Delta U[k]$, referred to as $\Delta W[k]$ hereafter:
\begin{eqnarray}
\Delta U[k] = \begin{bmatrix} \Delta V[k] \\ \Delta W[k] \end{bmatrix}
\label{eqn::simple_system_U_decompose}
\end{eqnarray}
where $\Delta V[k] \in \mathbb{R}^{M-P}$ and $\Delta W[k] \in \mathbb{R}^{P}$. Consider we decompose matrices $\tilde{Q}$ and $\tilde{R}$ as follows:
\begin{eqnarray}
\tilde{Q} = \begin{bmatrix} \tilde{Q}^{vv} & \tilde{Q}^{vw} \\ \tilde{Q}^{wv} & \tilde{Q}^{ww} \end{bmatrix}, \,\,\,\,
\tilde{R} = \begin{bmatrix} \tilde{R}^{vv} & \tilde{R}^{vw} \\ \tilde{R}^{wv} & \tilde{R}^{ww} \end{bmatrix} 
\label{eqn::simple_system_decompose_QR}
\end{eqnarray}
where the last lower right corner of size $P\times P$ is taken out with an index $^{ww}$. Likewise, system matrices  $\tilde{A}$ and $\tilde{B}$ can be decomposed to:
\begin{eqnarray}
\nonumber \begin{bmatrix} Y[k+1] \\ 0 \end{bmatrix} &=& 
\begin{bmatrix} \tilde{A}^{vv} & \tilde{A}^{vw} \\ \tilde{A}^{wv} & \tilde{A}^{ww} \end{bmatrix}
\begin{bmatrix} Y[k] \\ 0 \end{bmatrix} \\ 
&+& \begin{bmatrix} \tilde{B}^{vv} & \tilde{B}^{vw} \\ \tilde{B}^{wv} & \tilde{B}^{ww} \end{bmatrix}
\begin{bmatrix} \Delta V[k] \\ \Delta W[k] \end{bmatrix}
\label{eqn::simple_system_decompose_AB}
\end{eqnarray}
With such decomposition, we can take $\Delta W[k]$ out of (\ref{eqn::simple_system_decompose_AB}) from the last $P$ rows:
\begin{eqnarray}
\nonumber \Delta W[k] &=& \tilde{G}Y[k] + \tilde{H}\Delta V[k] \\
\nonumber \tilde{G} &=& - (\tilde{B}^{ww})^{-1}\tilde{A}^{wv} \\
\tilde{H} &=& - (\tilde{B}^{ww})^{-1}\tilde{B}^{wv}
\label{eqn::simple_system_W_GH}
\end{eqnarray}
and:
\begin{eqnarray}
\nonumber  Y[k+1] &=& \bar{A}Y[k]+\bar{B}\Delta V[k] \\
\nonumber \bar{A} &=& \tilde{A}^{vv} + \tilde{B}^{vw}  \tilde{G}\\
\bar{B} &=& \tilde{B}^{vv} + \tilde{B}^{vw}  \tilde{H}
\label{eqn::simple_system_lqr_free_appA_system_matrices}
\end{eqnarray}
Now, given that we have resolved the constraint, we can form an equivalent DLQR design in terms of $Y[k]$ and $\Delta V[k]$ by replacing $\Delta W[k]$ in all terms:
\begin{eqnarray}
\nonumber & \underset{Y[k],\Delta V[k]}{\text{min}} \sum_{k=0}^{\infty} \\
\nonumber & Y[k]^T\bar{Q}Y[k]+\Delta V[k]^T\bar{R}\Delta V[k]+2Y[k]^T\bar{N}\Delta V[k] \\
& s.t. \, \, \, Y[k+1] = \bar{A}Y[k]+\bar{B}\Delta V[k]
\, \, \, \, k\ge 0
\label{eqn::simple_system_lqr_free_appA}
\end{eqnarray}
which has a standard DLQR format without constraint. The new cost matrices in (\ref{eqn::simple_system_lqr_free_appA}) are defined as:
\begin{eqnarray}
\nonumber \bar{Q} &=& \tilde{Q}^{vv} + \tilde{G}^T \tilde{R}^{ww} \tilde{G}\\
\nonumber \bar{R} &=& \tilde{R}^{vv} + \tilde{H}^T \tilde{R}^{ww} \tilde{H} + \tilde{R}^{vw} \tilde{H} + \tilde{H}^T \tilde{R}^{wv}\\
\bar{N} &=& \tilde{G}^T ( \tilde{R}^{ww} \tilde{H} + \tilde{R}^{ww}{}^T \tilde{H}^T + \tilde{R}^{vw}{}^T +  \tilde{R}^{wv})
\label{eqn::simple_system_lqr_free_appA_matrices}
\end{eqnarray}
We call the optimal DLQR gain matrix for (\ref{eqn::simple_system_lqr_free_appA}) $\bar{K}$ which acts on $Y[k]=\tilde{C}E[k]$ and produces $\Delta V[k]$. The other part of system input $\Delta W[k]$ can be calculated by (\ref{eqn::simple_system_W_GH}). Overall, the system input is: 
\begin{eqnarray}
\Delta U[k] = \begin{bmatrix} -\bar{K} \\ \tilde{G}-\tilde{H}\bar{K} \end{bmatrix} \tilde{C}E[k]
\label{eqn::simple_system_lqr_free_sol}
\end{eqnarray}
which optimally satisfies the constraint and initial DLQR problem in (\ref{eqn::simple_system_lqr_appA}).

\subsection{Appendix B: Time-projection for constrained systems}
Time-projection controller for a constrained system is derived very similarly to a normal system. Consider all formulations and decompositions of Appendix A. The instantaneous error $e(t)=x(t)-\bar{x}(t)$ in a constrained system would evolve until the next control time by:
\begin{eqnarray}
\hat{E}_t[k+1] = A(\tau) e(t) + B(\tau) \delta \hat{U}_t[k]
\label{eqn::simple_system_xtoX+_appB}
\end{eqnarray}
where $\tau=(k+1)T-t$ and the constraint applies $C\hat{E}_t[k+1] = 0$. Here we assume a constant input $\delta \hat{U}_t[k]$ applied to the system which yields to a predicted error $\hat{E}_t[k+1]$. Remember the hat notation is used to emphasize that these predicted quantities are just calculated at time $t$ and they are not real system variables. Imagine we define $z(t) = Se(t)$, $\tilde{A}_t = S A(\tau) S^{-1}$, $\tilde{B}_t = S B(\tau)$ and perform a decomposition similar to (\ref{eqn::simple_system_decompose_AB}):
\begin{eqnarray}
\nonumber \hat{Z}_t[k+1] &=& \tilde{A}_t z(t) + \tilde{B}_t  \delta \hat{U}_t[k] \\
\nonumber \begin{bmatrix} \hat{Y}_t[k+1] \\ 0 \end{bmatrix} &=& 
\begin{bmatrix} \tilde{A}_t^{v} \\ \tilde{A}_t^{w} \end{bmatrix}
z(t) \\ 
&+& \begin{bmatrix} \tilde{B}_t^{vv} & \tilde{B}_t^{vw} \\ \tilde{B}_t^{wv} & \tilde{B}_t^{ww} \end{bmatrix}
\begin{bmatrix} \delta \hat{V}_t[k] \\ \delta \hat{W}_t[k] \end{bmatrix}
\label{eqn::simple_system_decompose_AcBc}
\end{eqnarray}
which describes system evolution from time $t$ to $(k+1)T$ whereas the constraint in (\ref{eqn::simple_system_lqr_Z_optim}) describes same evolution from $kT$ to $(k+1)T$. Here the subscript $t$ indicates dependency on $t$. Note that the last $P$ elements of $z(t)$ might not be zero in $kT<t<(k+1)T$, but the constraint implies that these last $P$ elements have to be zero at time instances $kT$ and $(k+1)T$. Similar to (\ref{eqn::simple_system_W_GH}), we can take $\delta \hat{W}[k]$ out of the last $P$ equations:
\begin{eqnarray}
\nonumber \delta \hat{W}_t[k] &=& \tilde{G}_t z(t) + \tilde{H}_t \delta \hat{V}_t[k] \\
\nonumber \tilde{G}_t &=& - ({\tilde{B}_t^{ww}})^{-1}\tilde{A}_t^{w} \\
\tilde{H}_t &=& - ({\tilde{B}_t^{ww}})^{-1}\tilde{B}_t^{wv}
\label{eqn::simple_system_W_GcHc}
\end{eqnarray}
and new system matrices would be defined as:
\begin{eqnarray}
\nonumber \hat{Y}_t[k+1] &=& \bar{A}_t z(t) + \bar{B}_t \delta \hat{V}_t[k] \\
\nonumber \bar{A}_t &=& \tilde{A}_t^{v} + \tilde{B}_t^{vw}  \tilde{G}_t\\
\bar{B}_t &=& \tilde{B}_t^{vv} + \tilde{B}_t^{vw}  \tilde{H}_t
\label{eqn::simple_system_lqr_free_AcbarBcbar}
\end{eqnarray}
Note that in (\ref{eqn::simple_system_lqr_free_appA_system_matrices}) and (\ref{eqn::simple_system_lqr_free_AcbarBcbar}), the constraint is resolved. In other words, system matrices are adjusted such that they account for the effect of $W$ inputs which aim at satisfying the constraint. Now, we can consider time-projection for this free system as follows. Imagine an initial reduced state $\hat{Y}_t[k]$ evolves in time by $\delta \hat{V}_t[k]$ according to  (\ref{eqn::simple_system_lqr_free_appA_system_matrices}) and yields to $\hat{Y}_t[k+1]$. Similarly, the current error $z(t)$ evolves in time by $\delta \hat{V}_t[k]$ and yields to the same $\hat{Y}_t[k+1]$ according to (\ref{eqn::simple_system_lqr_free_AcbarBcbar}).  Now, the system of equations in (\ref{eqn::simple_system_solvedU}) can be applied here as well:
\begin{eqnarray}
\begin{bmatrix} \bar{A} & \bar{B}-\bar{B}_t & \cdot\\ K & I & \cdot \\ \cdot & -\tilde{H}_t & I \end{bmatrix} \begin{bmatrix}
\hat{Y}_t[k] \\ \delta \hat{V}_t[k] \\ \delta \hat{W}_t[k] \end{bmatrix} = 
\begin{bmatrix} \bar{A}_t \\ 0 \\ \tilde{G}_t \end{bmatrix} S e(t)
\label{eqn::simple_system_solvedUc}
\end{eqnarray}
where the solution defines $\delta u(t) = \delta \hat{U}_t[k]$ to be applied at time instance $t$.

\subsection{Appendix C: Time-projection for a simple system}
\begin{figure}[]
	\centering
	\includegraphics[trim = 0mm 0mm 0mm 0mm, clip, width=0.5\textwidth]{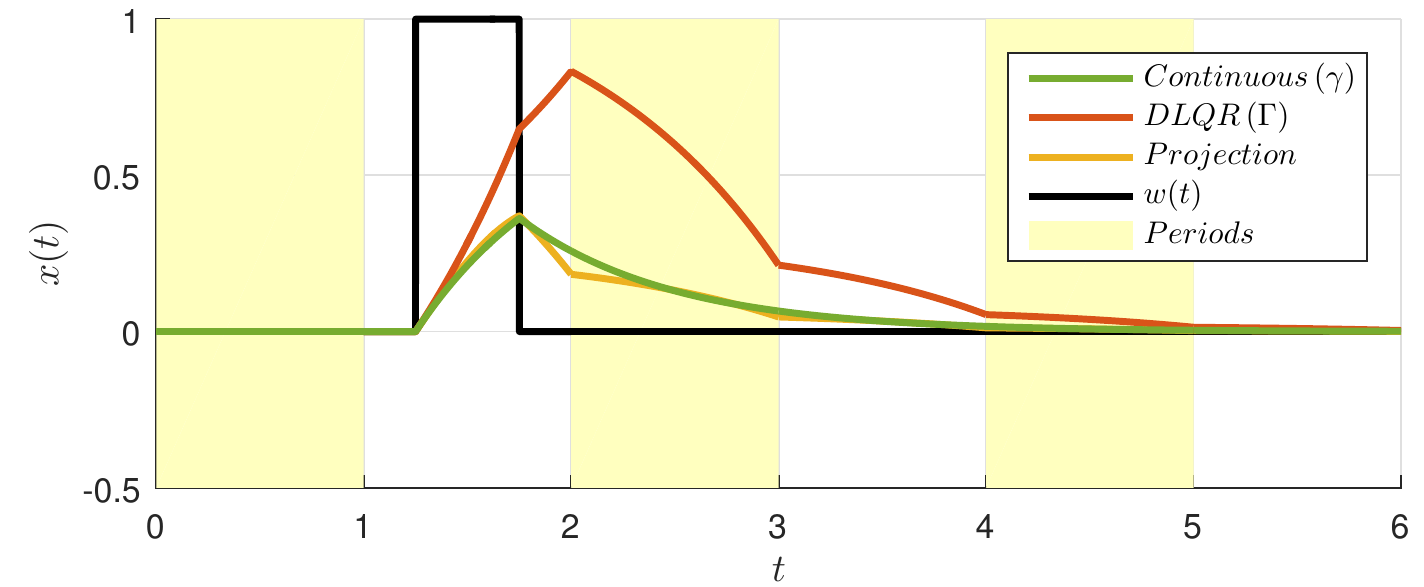}
	\caption{One degree-of-freedom system with continuous, DLQR and time-projection controllers.} 
	\label{fig::1dof}
\end{figure}

Consider a simple system of:
\begin{eqnarray}
\dot{x}(t) = x(t) + u(t) + w(t)
\label{eqn::1dof_system}
\end{eqnarray}
where $u(t)$ is the control input and $w(t)$ is disturbance. Consider a control period $T$ and steady-state solutions $\bar{x}=0$ and $\bar{u}=0$. The closed-form evolution of this system could be written as:
\begin{eqnarray}
X[k+1] = e^TX[k]+(e^T-1)U[k]
\label{eqn::1dof_system_discrete}
\end{eqnarray}
assuming a constant $U[k]$ applied to the system. Now, imagine we use a discrete controller at time instances $kT$ to adjust the constant input with a law of:
\begin{eqnarray}
U[k] = -\Gamma X[k]
\label{eqn::1dof_system_lqr}
\end{eqnarray}
The stability of closed-loop system suggests that:
\begin{eqnarray}
|e^T-\Gamma (e^T-1)|<1
\label{eqn::1dof_system_lqr_stability}
\end{eqnarray}
which puts boundaries on $\Gamma$:
\begin{eqnarray}
1<\Gamma<\frac{e^T+1}{e^T-1}
\label{eqn::1dof_system_lqr_Gamma_bound}
\end{eqnarray}
For this system, the DLQR controller satisfies this criteria. The time-projection controller takes a state $x(t)$, maps it to the previous time-sample $kT$ and finds the control input based on the discrete controller $\Gamma$. Without loss of generality, assume $t$ is between the first two time-samples ($kT<t<(k+1)T$) and:
\begin{eqnarray}
\nonumber x(t) &=& e^t \hat{X}_t[k] + (e^t-1)\delta \hat{U}_t[k]\\
\delta \hat{U}_t[k] &=& - \Gamma \hat{X}_t[k]
\label{eqn::1dof_system_projection}
\end{eqnarray}
where $\hat{X}_t[k]$ is a possible predicted initial state that can lead the current measurement $x(t)$. The two laws of (\ref{eqn::1dof_system_projection}) and system equations (\ref{eqn::1dof_system}) result in the following closed-loop system:
\begin{eqnarray}
\nonumber x(t) &=& e^t (-\frac{\delta \hat{U}_t[k]}{\Gamma}) + (e^t-1) \delta \hat{U}_t[k] \\
\dot{x}(t) &=& (1+\frac{1}{-\frac{e^t}{\Gamma} + e^t-1}) x(t)
\label{eqn::1dof_system_projection_closed_loop}
\end{eqnarray}
which is found by resolving $\hat{X}_t[k]$ in (\ref{eqn::1dof_system_projection}) and plugging $u(t) =\delta \hat{U}_t[k]$ in (\ref{eqn::1dof_system}). In order to produce finite feedbacks, the denominator in (\ref{eqn::1dof_system_projection_closed_loop}) should not have a zero in $0<t<T$. In other words, the root of denominator $t_0$ should be outside $[0,T]$.
\begin{eqnarray}
-\frac{e^{t_0}}{\Gamma} + e^{t_0}-1 = 0 \quad \rightarrow \quad t_0 = ln(\frac{1}{1-\frac{1}{\Gamma}})
\label{eqn::1dof_system_projection_root}
\end{eqnarray}
This leads to the following two conditions:
\begin{eqnarray}
\left\{ \begin{array}{l} ln(\frac{1}{1-\frac{1}{\Gamma}})<0 \rightarrow \frac{1}{1-\frac{1}{\Gamma}}<1 \rightarrow \Gamma<0 \\ ln(\frac{1}{1-\frac{1}{\Gamma}})>T  \rightarrow \frac{1}{1-\frac{1}{\Gamma}} > e^T \rightarrow \Gamma<\frac{e^T}{e^T-1} \end{array} \right. 
\label{eqn::1dof_system_projection_root_infinite}
\end{eqnarray}
This will further tighten the boundaries of (\ref{eqn::1dof_system_lqr_Gamma_bound}) to:
\begin{eqnarray}
1<\Gamma<\frac{e^T}{e^T-1}
\label{eqn::1dof_system_lqr_Gamma_newbound}
\end{eqnarray}
Note that DLQR design does not necessarily satisfy this criteria, unless proper state and input cost matrices are chosen in the objective. Consider the lyapunov function $V(t) = \frac{1}{2} x(t)^2$ whose derivative is:
\begin{eqnarray}
\dot{V}(t) = x(t) \dot{x}(t) = \epsilon (t) x(t)^2
\label{eqn::1dof_system_dlyap}
\end{eqnarray}
where:
\begin{eqnarray}
\epsilon (t) = 1+\frac{1}{-\frac{e^t}{\Gamma} + e^t-1}
\label{eqn::1dof_system_eps}
\end{eqnarray}
It is obvious that $\epsilon(t)$ is monotonically decreasing, because $\Gamma>1$ and thus:
\begin{eqnarray}
\dot{\epsilon} (t) = \frac{e^t(\frac{1}{\Gamma} - 1)}{(-\frac{e^t}{\Gamma} + e^t-1)^2} < 0
\label{eqn::1dof_system_deps}
\end{eqnarray}
Since $\epsilon(0) = 1-\Gamma < 0$, one can conclude that:
\begin{eqnarray}
\dot{V}(t) = \epsilon (t) x(t)^2 < \epsilon(0) x(t)^2
\label{eqn::1dof_system_lyapeps0}
\end{eqnarray}
which proves that $V(t)$ is decreasing and the system is stable with any choice of $\Gamma$ in (\ref{eqn::1dof_system_lqr_Gamma_newbound}). In order to compare the LQR and time-projection controllers with a simple continuous controller of $u(t) = -\gamma x(t)$, we find the discrete feedback gain $\Gamma$ by converting closed-loop eigen-values. The discrete eigen-value ($\Lambda$) of (\ref{eqn::1dof_system_lqr_stability}) is converted to a continuous eigen-value ($\lambda$) by a logarithm operation:
\begin{eqnarray}
\Lambda = e^T-\Gamma (e^T-1) = e^{\lambda T}
\label{eqn::1dof_system_discrete_eigen}
\end{eqnarray}
Given that $\lambda = 1 - \gamma$, we can $\lambda$ as:
\begin{eqnarray}
\gamma = -ln(e^T-\Gamma (e^T-1))/T+1
\label{eqn::1dof_system_continuous_eigen}
\end{eqnarray}
This let us compare the three types of controller on a fair basis. Consider we select $T=1$, $Q=1$ and $R=1$ for DLQR design which yields $\Gamma \simeq 1.43$ and it satisfies (\ref{eqn::1dof_system_lqr_Gamma_newbound}). The equivalent continuous gain for this system would be $\gamma \simeq 2.37$. Figure \ref{fig::1dof} demonstrates how an inter-sample disturbance could be rejected by these controllers. The DLQR controller overshoots, simply because of a late response which only starts at time $t=2$. The continuous controller $\gamma$ smoothly damps the disturbance while the time-projection controller performs damping in multiple steps with the same rate as the continuous controller. One can imagine that with decreasing $T$, the time-projection controller will converge to the continuous controller. 

\bibliographystyle{SageH}
\bibliography{Biblio}

\begin{thebibliography}{34}
\providecommand{\natexlab}[1]{#1}
\providecommand{\url}[1]{\texttt{#1}}
\providecommand{\urlprefix}{URL }
\expandafter\ifx\csname urlstyle\endcsname\relax
  \providecommand{\doi}[1]{DOI:\discretionary{}{}{}#1}\else
  \providecommand{\doi}{DOI:\discretionary{}{}{}\begingroup
  \urlstyle{rm}\Url}\fi

\bibitem[{Asano et~al.(2004)Asano, Yamakita, Kamamichi and
  Luo}]{asano2004novel}
Asano F, Yamakita M, Kamamichi N and Luo ZW (2004) A novel gait generation for
  biped walking robots based on mechanical energy constraint.
\newblock \emph{Robotics and Automation, IEEE Transactions on} 20(3): 565--573.

\bibitem[{Bertram(2005)}]{bertram2005constrained}
Bertram JE (2005) Constrained optimization in human walking: cost minimization
  and gait plasticity.
\newblock \emph{Journal of experimental biology} 208(6): 979--991.

\bibitem[{Bhounsule et~al.(2015)Bhounsule, Ruina and
  Stiesberg}]{bhounsule2015discrete}
Bhounsule PA, Ruina A and Stiesberg G (2015) Discrete-decision
  continuous-actuation control: balance of an inverted pendulum and pumping a
  pendulum swing.
\newblock \emph{Journal of Dynamic Systems, Measurement, and Control} 137(5):
  051012.

\bibitem[{Byl and Tedrake(2008)}]{byl2008approximate}
Byl K and Tedrake R (2008) Approximate optimal control of the compass gait on
  rough terrain.
\newblock In: \emph{Robotics and Automation, 2008. ICRA 2008. IEEE
  International Conference on}. IEEE, pp. 1258--1263.

\bibitem[{De~Leva(1996)}]{de1996adjustments}
De~Leva P (1996) Adjustments to zatsiorsky-seluyanov's segment inertia
  parameters.
\newblock \emph{Journal of biomechanics} 29(9): 1223--1230.

\bibitem[{Faraji et~al.(2015)Faraji, Colasanto and
  Ijspeert}]{faraji2015practical}
Faraji S, Colasanto L and Ijspeert AJ (2015) Practical considerations in using
  inverse dynamics on a humanoid robot: Torque tracking, sensor fusion and
  cartesian control laws.
\newblock In: \emph{Intelligent Robots and Systems (IROS), 2015 IEEE/RSJ
  International Conference on}. IEEE, pp. 1619--1626.

\bibitem[{Faraji and Ijspeert(2017)}]{faraji20173lp}
Faraji S and Ijspeert AJ (2017) 3lp: A linear 3d-walking model including torso
  and swing dynamics.
\newblock \emph{the international journal of robotics research} 36(4):
  436--455.

\bibitem[{Faraji et~al.(2014)Faraji, Pouya and Ijspeert}]{faraji2014robust}
Faraji S, Pouya S and Ijspeert A (2014) Robust and agile 3d biped walking with
  steering capability using a footstep predictive approach.
\newblock In: \emph{Robotics Science and Systems (RSS)}.

\bibitem[{Feng et~al.(2014)Feng, Whitman, Xinjilefu and
  Atkeson}]{Feng2014optimization}
Feng S, Whitman E, Xinjilefu X and Atkeson CG (2014) Optimization-based full
  body control for the darpa robotics challenge .

\bibitem[{Feng et~al.(2013)Feng, Xinjilefu, Huang and Atkeson}]{feng20133d}
Feng S, Xinjilefu X, Huang W and Atkeson CG (2013) 3d walking based on online
  optimization.
\newblock In: \emph{Humanoid Robots (Humanoids), 2013 13th IEEE-RAS
  International Conference on}. IEEE, pp. 21--27.

\bibitem[{Gawthrop et~al.(2011)Gawthrop, Loram, Lakie and
  Gollee}]{gawthrop2011intermittent}
Gawthrop P, Loram I, Lakie M and Gollee H (2011) Intermittent control: a
  computational theory of human control.
\newblock \emph{Biological cybernetics} 104(1-2): 31--51.

\bibitem[{Gawthrop and Wang(2007)}]{gawthrop2007intermittent}
Gawthrop PJ and Wang L (2007) Intermittent model predictive control.
\newblock \emph{Proceedings of the Institution of Mechanical Engineers, Part I:
  Journal of Systems and Control Engineering} 221(7): 1007--1018.

\bibitem[{Geyer and Herr(2010)}]{geyer2010muscle}
Geyer H and Herr H (2010) A muscle-reflex model that encodes principles of
  legged mechanics produces human walking dynamics and muscle activities.
\newblock \emph{Neural Systems and Rehabilitation Engineering, IEEE
  Transactions on} 18(3): 263--273.

\bibitem[{Gregg et~al.(2012)Gregg, Tilton, Candido, Bretl and
  Spong}]{gregg2012control}
Gregg RD, Tilton AK, Candido S, Bretl T and Spong MW (2012) Control and
  planning of 3-d dynamic walking with asymptotically stable gait primitives.
\newblock \emph{Robotics, IEEE Transactions on} 28(6): 1415--1423.

\bibitem[{Herdt et~al.(2010{\natexlab{a}})Herdt, Diedam, Wieber, Dimitrov,
  Mombaur and Diehl}]{herdt2010online}
Herdt A, Diedam H, Wieber PB, Dimitrov D, Mombaur K and Diehl M
  (2010{\natexlab{a}}) Online walking motion generation with automatic footstep
  placement.
\newblock \emph{Advanced Robotics} 24(5-6): 719--737.

\bibitem[{Herdt et~al.(2010{\natexlab{b}})Herdt, Perrin and
  Wieber}]{herdt2010walking}
Herdt A, Perrin N and Wieber PB (2010{\natexlab{b}}) Walking without thinking
  about it.
\newblock In: \emph{Intelligent Robots and Systems (IROS), 2010 IEEE/RSJ
  International Conference on}. IEEE, pp. 190--195.

\bibitem[{Kajita and Tani(1991)}]{kajita1991study}
Kajita S and Tani K (1991) Study of dynamic biped locomotion on rugged
  terrain-derivation and application of the linear inverted pendulum mode.
\newblock In: \emph{Robotics and Automation, 1991. Proceedings., 1991 IEEE
  International Conference on}. IEEE, pp. 1405--1411.

\bibitem[{Kelly and Ruina(2015)}]{kelly2015non}
Kelly M and Ruina A (2015) Non-linear robust control for inverted-pendulum 2d
  walking.
\newblock In: \emph{Robotics and Automation (ICRA), 2015 IEEE International
  Conference on}. IEEE, pp. 4353--4358.

\bibitem[{Koolen et~al.(2012)Koolen, De~Boer, Rebula, Goswami and
  Pratt}]{capturability}
Koolen T, De~Boer T, Rebula J, Goswami A and Pratt J (2012) Capturability-based
  analysis and control of legged locomotion, part 1: Theory and application to
  three simple gait models.
\newblock \emph{The International Journal of Robotics Research} 31(9):
  1094--1113.

\bibitem[{Kuindersma et~al.(2014)Kuindersma, Permenter and
  Tedrake}]{kuindersma2014efficiently}
Kuindersma S, Permenter F and Tedrake R (2014) An efficiently solvable
  quadratic program for stabilizing dynamic locomotion.
\newblock In: \emph{Robotics and Automation (ICRA), 2014 IEEE International
  Conference on}. IEEE, pp. 2589--2594.

\bibitem[{Kuo et~al.(2005)Kuo, Donelan and Ruina}]{kuo2005energetic}
Kuo AD, Donelan JM and Ruina A (2005) Energetic consequences of walking like an
  inverted pendulum: step-to-step transitions.
\newblock \emph{Exercise and sport sciences reviews} 33(2): 88--97.

\bibitem[{Manchester and Umenberger(2014)}]{manchester2014real}
Manchester IR and Umenberger J (2014) Real-time planning with primitives for
  dynamic walking over uneven terrain.
\newblock In: \emph{Robotics and Automation (ICRA), 2014 IEEE International
  Conference on}. IEEE, pp. 4639--4646.

\bibitem[{Moro et~al.(2011)Moro, Tsagarakis and Caldwell}]{coman}
Moro FL, Tsagarakis NG and Caldwell DG (2011) A human-like walking for the
  compliant humanoid coman based on com trajectory reconstruction from
  kinematic motion primitives.
\newblock In: \emph{Humanoid Robots (Humanoids), 2011 11th IEEE-RAS
  International Conference on}. IEEE, pp. 364--370.

\bibitem[{Ogata(1995)}]{ogata1995discrete}
Ogata K (1995) \emph{Discrete-time control systems}, volume~2.
\newblock Prentice Hall Englewood Cliffs, NJ.

\bibitem[{Pratt et~al.(2012)Pratt, Koolen, De~Boer, Rebula, Cotton, Carff,
  Johnson and Neuhaus}]{capturability2}
Pratt J, Koolen T, De~Boer T, Rebula J, Cotton S, Carff J, Johnson M and
  Neuhaus P (2012) Capturability-based analysis and control of legged
  locomotion, part 2: Application to {M2V2}, a lower-body humanoid.
\newblock \emph{The International Journal of Robotics Research} 31(10):
  1117--1133.

\bibitem[{Raibert et~al.(1984)Raibert, Brown and
  Chepponis}]{raibert1984experiments}
Raibert MH, Brown HB and Chepponis M (1984) Experiments in balance with a 3d
  one-legged hopping machine.
\newblock \emph{The International Journal of Robotics Research} 3(2): 75--92.

\bibitem[{Rummel et~al.(2010)Rummel, Blum, Maus, Rode and
  Seyfarth}]{rummel2010stable}
Rummel J, Blum Y, Maus HM, Rode C and Seyfarth A (2010) Stable and robust
  walking with compliant legs.
\newblock In: \emph{Robotics and Automation (ICRA), 2010 IEEE International
  Conference on}. IEEE, pp. 5250--5255.

\bibitem[{Sharbafi et~al.(2012)Sharbafi, Maufroy, Maus, Seyfarth, Ahmadabadi
  and Yazdanpanah}]{sharbafi2012controllers}
Sharbafi MA, Maufroy C, Maus HM, Seyfarth A, Ahmadabadi MN and Yazdanpanah MJ
  (2012) Controllers for robust hopping with upright trunk based on the virtual
  pendulum concept.
\newblock In: \emph{Intelligent Robots and Systems (IROS), 2012 IEEE/RSJ
  International Conference on}. IEEE, pp. 2222--2227.

\bibitem[{Stephens(2007)}]{stephens2007humanoid}
Stephens B (2007) Humanoid push recovery.
\newblock In: \emph{Humanoid Robots, 2007 7th IEEE-RAS International Conference
  on}. IEEE, pp. 589--595.

\bibitem[{Teschl(2012)}]{poincare}
Teschl G (2012) \emph{Ordinary differential equations and dynamical systems},
  volume 140.
\newblock American Mathematical Soc.

\bibitem[{Tsagarakis et~al.(2017)Tsagarakis, Caldwell, Negrello, Choi,
  Baccelliere, Loc, Noorden, Muratore, Margan, Cardellino, Natale,
  Mingo~Hoffman, Dallali, Kashiri, Malzahn, Lee, Kryczka, Kanoulas, Garabini,
  Catalano, Ferrati, Varricchio, Pallottino, Pavan, Bicchi, Settimi, Rocchi and
  Ajoudani}]{tsagarakis2017walk}
Tsagarakis NG, Caldwell DG, Negrello F, Choi W, Baccelliere L, Loc V, Noorden
  J, Muratore L, Margan A, Cardellino A, Natale L, Mingo~Hoffman E, Dallali H,
  Kashiri N, Malzahn J, Lee J, Kryczka P, Kanoulas D, Garabini M, Catalano M,
  Ferrati M, Varricchio V, Pallottino L, Pavan C, Bicchi A, Settimi A, Rocchi A
  and Ajoudani A (2017) Walk-man: A high-performance humanoid platform for
  realistic environments.
\newblock \emph{Journal of Field Robotics} 34(7): 1225--1259.
\newblock \doi{10.1002/rob.21702}.

\bibitem[{Westervelt et~al.(2007)Westervelt, Grizzle, Chevallereau, Choi and
  Morris}]{westervelt2007feedback}
Westervelt ER, Grizzle JW, Chevallereau C, Choi JH and Morris B (2007)
  \emph{Feedback control of dynamic bipedal robot locomotion}, volume~28.
\newblock CRC press.

\bibitem[{Zarrugh and Radcliffe(1978)}]{zarrugh1978predicting}
Zarrugh M and Radcliffe C (1978) Predicting metabolic cost of level walking.
\newblock \emph{European Journal of Applied Physiology and Occupational
  Physiology} 38(3): 215--223.

\bibitem[{Zaytsev et~al.(2015)Zaytsev, Hasaneini and Ruina}]{zaytsev2015two}
Zaytsev P, Hasaneini S and Ruina A (2015) Two steps is enough: No need to plan
  far ahead for walking balance.
\newblock In: \emph{Robotics and Automation (ICRA), 2015 IEEE International
  Conference on}. pp. 6295--6300.
\newblock \doi{10.1109/ICRA.2015.7140083}.

\end{thebibliography}

\end{document}